\documentclass{amsart}
\usepackage[utf8]{inputenc}
\usepackage{amsmath, amssymb, amsthm,amsfonts,color}
\usepackage{hyperref}  
\usepackage{parskip}
\usepackage{verbatim}
\usepackage{yhmath}
\usepackage{enumerate}
\usepackage{csquotes}
\usepackage{chngcntr}

\usepackage{epsfig,mathrsfs,graphicx,euscript,amsxtra}

\newtheorem{theorem}{Theorem}

\newtheorem{lemma}[theorem]{Lemma}

\newtheorem{definition}[theorem]{Definition}

\theoremstyle{remark}

\numberwithin{equation}{section}

\newcommand{\Dec}{\text{Dec}}
\newcommand{\Enc}{\text{Enc}}
\newcommand{\Star}{\text{STAR}^\beta(C)}
\newcommand{\ave}{\text{Ave}}
\newcommand{\quotes}[1]{``#1''}
\numberwithin{theorem}{section} 
\usepackage{etoolbox}
\makeatletter
\patchcmd{\@maketitle}
  {\ifx\@empty\@dedicatory}
  {\ifx\@empty\@date \else {\vskip3ex \centering\footnotesize\@date\par\vskip1ex}\fi
   \ifx\@empty\@dedicatory}
  {}{}
\patchcmd{\@adminfootnotes}
  {\ifx\@empty\@date\else \@footnotetext{\@setdate}\fi}
  {}{}{}
\makeatother

\title{Optimal Approximation with Sparse Neural Networks and Applications}
\author{Hong Khay Boon}
\date{July 2021}

\begin{document}

\begin{abstract}
We use deep sparsely connected neural networks to measure the complexity of a function class in $L^2(\mathbb R^d)$ by restricting connectivity and memory requirement for storing the neural networks. We also introduce representation system - a countable collection of functions to guide neural networks, since approximation theory with representation system has been well developed in Mathematics. 
We then prove the fundamental bound theorem, implying a quantity intrinsic to the function class itself can give information about the approximation ability of neural networks and representation system. We also provides a method for transferring existing theories about approximation by representation systems to that of neural networks, greatly amplifying the practical values of neural networks.
Finally, we use neural networks to approximate B-spline functions, which are used to generate the B-spline curves. Then, we analyse the complexity of a class called $\beta$ cartoon-like functions using rate-distortion theory and wedgelets construction.

\smallskip
\keywordsname: Neural Networks, Effective Approximation Rate, Polynomial Depth-search, Sparsity, Rate-distortion Theory, Wedgelets, Edgelets.  
\end{abstract}

\maketitle

\section{Introduction}
Neural network, as a branch of machine learning, has been flexing its ability to our modern world. 
It helps us deal with ordinary daily tasks or collecting data for statistical purposes, which are mostly mundane tasks. 
Some examples are image classification \cite{img_clf}, speech to text conversion \cite{speech} and vehicle license plate recognition \cite{lpr}. 
Experts had been developing  theories such as back-propagation to equip neural networks with the ability of self-learning, even before powerful computers exist (See \cite{seppo}). 
The most common neural network learning processes are based on empirical evidence, from the simplest stochastic gradient descent to the most effective Adaptive Moment Estimation (Adam) algorithm \cite{adam}. 
Their performances are compared and concluded based on the datasets in Modified National Institute of Standards and Technology (MNIST) or the ImageNet Large Scale Visual Recognition Challenge (ILSVRC) \cite{mobilenetv2} \cite{ssd}. 
These learning algorithms have been put into practical uses despite not having the most rigorous mathematical background. 
However, Wu et. al. did showed a deterministic convergence of stochastic gradient descent with some restriction on the activation function \cite{det_sto}. 

In contrast to the practical value of neural networks, this paper mainly focus on its theoretical aspect \cite{main}. 
Given a function class, we use neural networks to approximate its functions. 
We could consider many possible settings, such as the number of layers, number of weights, whether the weights are discrete or the weights are bounded above. 
In order to focus on sparsity of neural networks, we will restrict the number of weights during approximation. 
Moreover, the weights are assumed to be bounded above and discretized, so that they can be stored in the computer and improve memory efficiency.
We focus on the theoretical limitation of the neural networks and derive important results. These include the fundamental bound theorem of the function class, and representability of certain representation system by neural networks. 

The main contribution of this paper is to examine the theoretical properties of two practical examples.
We use neural networks to approximate B-splines and carry out the derivation of optimal exponent of the $\beta$ cartoon-like function class. 
Since we have to write these proofs in great detail, we direct the reader to the original papers \sloppy{\cite{main}, \cite{donoho2} and \cite{donoho3}} in case the proof of the other theorems are not given. 
If the proof is given, it is either because we have original result as part of the proof, or more details that are not from the original papers are provided.

\section{Notations}
We introduce the definition of neural networks.
A standard neural network should have an input layer of $d$ nodes, followed by $L$ subsequent layers. 
Each layer has some nodes, and the connections are only made between consecutive layers. The connections are the edge weights, while each node itself possess a node weight. 
We will first define affine linear maps which record the edge and node weights of a neural network.

\begin{definition}\textit{Affine linear map}. A mapping $W:\mathbb R^n\to \mathbb R^m$ is an \emph{affine linear map} if there is $m\times n$ matrix $A$ and $m\times 1$ vector $b$ such that 
$$W(x)=Ax+b\quad\text{ for all }x\in\mathbb R^n.$$
The matrix $A$ represent edge weights, $b$ represent node weights.
\end{definition}

\begin{definition}
\textit{Deep Neural Network}. Let $L,d,N_1,\ldots, N_L$ be positive integers with $L\geq 2$. A map $\Phi:\mathbb R^d\to \mathbb R^{N_L}$ given by 
$$\Phi(x)=W_L\rho (W_{L-1}\rho(\ldots\rho(W_1(x))))\quad\text{ for }x\in\mathbb R^d,$$
with affine linear maps $W_\ell: \mathbb R^{N_{\ell-1}}\to \mathbb R^{N_\ell}, 1\leq \ell\leq L$, and the nonlinear activation function $\rho$ acting componentwise, is called a \emph{neural network}.
\end{definition}

If the activation function $\rho$ is linear (hence affine linear), then the neural network would equivalent to a single layer neural network.
It is because the composition of affine linear maps is an affine linear map. 
Note that in a neural network of $L$ layers, only $L-1$ activation functions were used. Therefore, we can serially concatenating two neural networks without needing an extra layer.

In the general discussion, the total number of nodes in the neural network is denoted as
$$\mathcal N(\Phi)=d+\sum_{\ell=1}^L N_\ell.$$
For the affine linear maps $W_\ell(x)=A_\ell x+b_\ell$ involved in the neural network, we define $\mathcal M(\Phi)$ to be the number of nonzero entries in all $A_\ell$ and $b_\ell$ for $\ell=1,2,\dots,L$. This quantity is also called the \textbf{connectivity} of neural network. We should always assume the neural network is a real-valued function, that is $N_L=1$. The results in this paper can be appropriately generalized to $N_L>1$, but we will not further elaborate that. 

Finally, we define the class of neural networks, which focus on the restriction on connectivity as well as the structure of neural networks.

\begin{definition}
\textit{Class of Neural Networks}. We let $\mathcal{NN}_{L,M,d,\rho}$  denote the collection of all neural networks $\Phi:\mathbb R^d\to \mathbb R$ with exactly $L$ layers, connectivity $\mathcal M(\Phi)$ no more than $M$, and activation function $\rho$. When $L=1$ the collection $\mathcal{NN}_{L,M,d,\rho}$ is empty. 

Moreover, we define 
\begin{align*}\mathcal{NN}_{\infty,M,d,\rho}&=\bigcup_{L\in\mathbb N}\mathcal{NN}_{L,M,d,\rho},\\
\mathcal{NN}_{L,\infty,d,\rho}&=\bigcup_{M\in\mathbb N}\mathcal{NN}_{L,M,d,\rho},\\
\mathcal{NN}_{\infty,\infty,d,\rho}&=\bigcup_{L\in\mathbb N}\mathcal{NN}_{L,\infty,d,\rho}.\end{align*}
\end{definition}
We will be mostly using $\mathcal{NN}_{\infty, M, d, \rho}$ in this paper, it is because we emphasize on the connectivity of neural networks rather than the number of layers.

\section{Evaluate Approximation Quality}
\subsection{Approximation by representation systems}
Before jumping into topics about approximation by neural networks, we should define the function class we are targeting. 
Given a Lebesgue measurable set $\Omega\subset \mathbb R^n$, we define $L^2(\Omega)$ to be the collection of all Lebesgue measurable functions $f:\Omega\to \mathbb R$ such that $$\int_\Omega f^2\text{ is finite}. $$
We give $L^2(\Omega)$ the usual metric topology $$\|f\|_{L^2(\Omega)} = \left(\int_\Omega f^2\right)^{1/2}$$
so $L^2(\Omega)$ is a Banach space. Then, we say $\mathcal C$ is a \textbf{function class} if it is a compact subset of $L^2(\Omega)$. A countable collection of functions $\mathcal D$ contained in $L^2(\Omega)$ is called a \textbf{representation system}. 

There are many approximation theories concerning approximation of a function class by a representation system. 
One familiar examples of approximation is when $\mathcal C=L([0,2\pi])$, the space of periodic Lebesgue-integrable functions on $[0,2\pi]$ with period $2\pi$, and when \sloppy{$\mathcal D=\{1, \sin x, \cos x, \sin 2x, \cos 2x,\ldots\}$}, the countable set of sinusoidal functions. 
The results from Fourier series showed that one can use finite linear combination of elements of $\mathcal D$ to approximate $\mathcal C$ arbitrarily well. 
Another example is when $\mathcal C=C([0,1])$, the space of continuous functions on $[0,1]$, and when $\mathcal D$ is the collection of monomials $\{1,x,x^2,\dots\}$.
The Bernstein polynomial can be used to approximation functions in $\mathcal C$ arbitrarily well. 
Moreover, when the function $f\in \mathcal C$ satisfies $f(0)=f(1)=0$, it can be approximated by polynomials with integer coefficients arbitrarily well \cite[p.~14]{conapprox}. 
Based on these existing theories, we could then transfer them to the study of approximation by neural networks. The method of doing so will be introduced in Chapter 6.

To quantify the quality of approximation, we study the \textbf{error of best $M$-term approximation of $f\in\mathcal C$ in $\mathcal D$}.

\begin{definition}
Given $d\in\mathbb N, \Omega\subset \mathbb R^d$, a function class $\mathcal C\subset L^2(\Omega)$, and a representation system $\mathcal D=(\varphi_i)_{i\in I}\subset L^2(\Omega)$, we define, for $f\in\mathcal C$ and $M\in \mathbb N$, 
\begin{equation}\label{bestM}\Gamma_M^{\mathcal D}(f):=\inf_{\substack{I_M\subset I, \\ \# I_M=M, (c_i)_{i\in I_M}}} \left\lVert f-\sum_{i\in I_M} c_i\varphi_i\right\rVert_{L^2(\Omega)}.\end{equation}
We call $\Gamma_M^{\mathcal D}(f)$ the \emph{best $M$-term approximation error} of $f$ in $\mathcal D$. If there exists $f_M=\sum_{i\in I_M} c_i\varphi_i$ attains the infimum in (\ref{bestM}), then $f_M$ is a best $M$-term approximation of $f$ in $\mathcal D$. 
\end{definition}

It is clear that choosing linear combinations of more terms from $\mathcal D$ will improve the approximation. We thus have 
$$\Gamma_M^{\mathcal D}(f)\leq \Gamma_N^{\mathcal D}(f)\quad\text{ if }\quad M\geq N.$$
Assume $\Gamma_M^{\mathcal D}(f)\to 0$ for $f\in \mathcal C$ as $M\to\infty$ for a moment. To determine the speed of convergence to zero, we introduce a positive real number $\gamma$ satisfying 
\begin{equation}\label{gamma}\sup_{f\in \mathcal C}\Gamma_M^{\mathcal D}(f)\in \mathcal O(M^{-\gamma})\quad\text{ as }M\to \infty.\end{equation}
The Big-oh notation $\mathcal O(g(\cdot))$ for nonnegative function $g$ is the set of functions $f$ such that $|f(x)|\leq Cg(x)$ for some $C>0$ and for $x$ sufficiently large. In other words, relation (\ref{gamma}) implies there is a positive constant $C$ such that 
$$\sup_{f\in\mathcal C}\Gamma_M^{\mathcal D}(f)\leq CM^{-\gamma}\quad\text{ for $M$ sufficiently large.}$$
Note that such positive $\gamma$ might not exists, which can happen if there exists $f\in \mathcal C$ such that $\Gamma_M^{\mathcal D}(f)\not\to 0$, then we could say $\gamma=0$ in this case. Regardless, we should always assume the existence of positive $\gamma$ in the later discussion. 

Finally, to pursue an optimal value of $\gamma$, we define the following.
\begin{definition}
Define the supremum of all $\gamma>0$ satisfying estimation (\ref{gamma}) to be $\gamma^*(\mathcal C, \mathcal D)$, called the \emph{best $M$-term approximation rate} of $\mathcal C$ in the representation system $\mathcal D$.
\end{definition}

\subsection{Approximation by neural networks}
There is an analogue definition of best $M$-term approximation error for neural networks. To achieve that, it is appropriate to treat each nonzero edge and node weight in a neural network as an element in a representation system. The following definition will be used.

\begin{definition}
Given $d\in\mathbb N, \Omega\subset \mathbb R^d$, a function class $\mathcal C\subset L^2(\Omega)$, and an activation function $\rho: \mathbb R\to \mathbb R$, we define, for $f\in\mathcal C$ and a positive integer $M$, 
\begin{equation}\label{bestMNN}\Gamma_M^{\mathcal N}(f):=\inf_{\Phi\in\mathcal {NN}_{\infty, M, d, \rho}} \lVert f-\Phi\rVert_{L^2(\Omega)}.
\end{equation}
We call $\Gamma_M^{\mathcal N}(f)$ the \emph{best $M$-edge approximation error} of $f$. 
\end{definition}

Similarly, we are interested in positive real numbers $\gamma$ such that 
\begin{equation}\label{gammaNN}\sup_{f\in\mathcal C}\Gamma_M^{\mathcal {N}}(f)\in\mathcal O(M^{-\gamma})\quad\text{ as }M\to \infty,\end{equation}
so we have

\begin{definition}
The supremum of all $\gamma$ satisfying estimate (\ref{gammaNN}) is denoted by $\gamma_{\mathcal {NN}}^*(\mathcal C,\rho)$, called the \emph{best $M$-edge approximation rate} of $\mathcal C$ by neural networks with activation function $\rho$. 
\end{definition}

In fact, these definitions will not be useful at discovering the complexity of a function class. 
It has been shown in \cite{donoho} and \cite{grohs} that every dense and countable representation system $\mathcal D\subset L^2(\Omega)$ achieves supremum rate $\gamma^*(\mathcal C, \mathcal D)=\infty$  for every function class $\mathcal C\subset L^2(\Omega)$. 
The meaning of infinite $\gamma^*(\mathcal C,\mathcal D)$ implies one can approximate elements in $\mathcal C$ arbitrarily well by finite linear combinations of elements in $\mathcal D$. 
An example is when $\mathcal C=C[0,1]$, the space of continuous functions on the unit interval treated as a subspace of $L^2[0,1]$. Consider the sinusoidal basis $B=\{1,\sin2\pi x, \cos 2\pi x, \dots,\sin 2\pi nx, \cos 2\pi nx, \dots\}$, then let $\mathcal D$ be the finite linear combination of elements in $B$ with coefficients being rational numbers. 
We observe that $\mathcal D$ is a countable set, and by the result of Fourier series, $\mathcal D$ is dense in $\mathcal C$ in $L^2$ norm. 
Therefore, for $f\in \mathcal C$ and given $\varepsilon>0$, we only need to choose one element $g\in \mathcal D$ such that $\|f-g\|_2<\varepsilon$. 

The example above shows that $\gamma^*(\mathcal C, \mathcal D)=\infty$ regardless the complexity of $\mathcal C$. 
This motivates us to define a new notation in order to capture the complexity of $\mathcal C$. At least, we should expect the following:
Given a fix representation system $\mathcal D$ and two function classes $\mathcal C_1, \mathcal C_2$ such that $\mathcal C_1$ is more complicated than $\mathcal C_2$ (for example, $\mathcal C_2\subset \mathcal C_1$), then the supremum rate of the pair $(\mathcal C_1,\mathcal D)$ should be larger than that of $(\mathcal C_2, \mathcal D)$. 
Moreover, we should expect when the function class is complicated enough, the supremum rate will become finite. 
We develop these notations in the next chapter.

\section{Effective Search}

Suppose we are performing approximation of a function class with a computer. We store a representation system in it (which could be done by define a function depending on the index $n\in\mathbb N$). 
Then, we want to implement a method to search for the best $M$-term approximation of a function $f$. 
This would pose a problem for the computer because it is impossible for it to search among infinitely many terms. 
Similar situation happens when we want to use a computer to approximate functions by neural networks, with the notation of best $M$-edge approximation error. 
Therefore, we will sacrifice some theoretical accuracy to makes it practically possible. 
For this reason, we call these practical version of best $M$-term approximation the \quotes{\textit{Effective} best $M$-term approximation}.

\subsection{Effective best M-term approximation}
To solve the problem of searching $M$ terms from all terms of $\mathcal D=\{\varphi_i\}_{i=1}^\infty$, we restrict the search on the first few terms. 
Ideally, let $\pi$ be a polynomial with integer coefficients, we only search the $M$ terms from $\{\varphi_1,\varphi_2,\dots,\varphi_{\pi(M)}\}$, called \textbf{polynomial search}. 
This makes it possible for a computer to search for candidates of approximation because it knows when to stop the search. 

\begin{definition}\label{bestMterm}
Given $d\in\mathbb N, \Omega\subset \mathbb R^d$, a function class $\mathcal C\subset L^2(\Omega)$, and a representation system $\mathcal D=(\varphi_i)_{i\in I}\subset L^2(\Omega)$, the supremum $\gamma>0$ so that there exist a polynomial $\pi$ and a constant $D>0$ such that 
\begin{align*}
    \sup_{f\in\mathcal C}\inf_{\substack{I_M\subset {1, \ldots,\pi(M)}\\ \# I_M=M, (c_i)_{i\in I_M}, \max_{i\in I_M}|c_i|\leq D}} \left\lVert f-\sum_{i\in I_M}c_i\varphi_i\right\rVert_{L^2(\Omega)}\in\mathcal O(M^{-\gamma})\text{ as }M\to \infty,
\end{align*}
will be denoted by $\gamma^{*, e}(\mathcal C,\mathcal D)$ and referred to as \emph{effective best $M$-term approximation rate} of $\mathcal C$ in the representation system $\mathcal D$. 
\end{definition}

Besides using polynomial depth-search, we also restrict the coefficients so that they are bounded above.
This could make the search more feasible. 
In the next chapter, we will observe that $\sup_{\mathcal D}\gamma^{*,e}(\mathcal C, \mathcal D)$ is bounded above by a quantity that only depends on the function class $\mathcal C$, where the supremum is taken across all representation systems in $L^2(\Omega)$.  

\subsection{Effective best M-edge approximation}

Similarly, it has been shown that any function $f\in C([0,1]^d)$ can be approximated arbitrarily well with a uniform error $\varepsilon>0$ by a three layer neural network with specific settings. 

If $\rho: \mathbb R\to \mathbb R$ is an activation function which is infinitely differentiable, strictly increasing, and $\lim_{x\to \infty} \rho(x)=1, \lim_{x\to -\infty} \rho(x)=0$, then we call this a \textbf{plain sigmoidal activation function}. 
Maiorov and Pinkus proved the following theorem: 

\begin{theorem}\cite{maiorov}
There exists a plain sigmoidal activation function $\rho$ such that for any $d\in\mathbb N$, $f\in C([0,1]^d)$ and $\varepsilon>0$, there is a neural network $\Phi\in\mathcal{NN}_{3, M, d, \rho}$ satisfying
$$\sup_{x\in[0,1]^d} |f(x)-\Phi(x)|\leq \varepsilon.$$
Moreover, the three layers after the input layer has $N_1=3d, N_2=6d+3, N_3=1$ nodes respectively, so the connectivity is $\mathcal M(\Phi)\leq M:=21d^2+15d+3$. 
\end{theorem}

Note that this result can be extended to $C(K)$, where $K$ is any compact $n$-dimensional interval of $\mathbb R^n$. 
It is surprising that we can consistently using neural networks of three layers to approximate continuous functions of compact support, it is even important to note that the networks can have no more than $21d^2+15d+3$ connectivity, resulting in an approximation obeying arbitrary uniform error.
Therefore, we have $\gamma_{\mathcal{NN}}^*(\mathcal C, \rho) = +\infty$.
However, as pointed out in \cite{main}, the edge and node weights cannot have their magnitude bounded above by $\mathcal O(\varepsilon^{-1})$.
Therefore, it is not reasonable to expect a computer to search for a neural network approximant because it can only worked with bounded weights.
For this reason, we define a class of neural network that puts more restrictions on its weights.

\begin{definition}\label{effectivenn}
Let $d, L, M\in \mathbb N$, $\rho:\mathbb R\to \mathbb R$ an activation function. Let $\pi$ be a polynomial,  then $\mathcal {NN}_{L,M,d,\rho}^\pi$ is the class of neural networks in $\mathcal {NN}_{L,M,d,\rho}$ where the weights are bounded above in absolute value by $|\pi(M)|$. 
\end{definition}

We don't restrict the weights to be bounded above by some fixed constant, but we allow the bound to grow polynomially as the connectivity. 
This relaxation will help us later.
Talking about storing information, suppose we know an integer is between $1$ and $N$ and we want to record it using a bitstring consists of 0s and 1s. 
Since a bitstring of length $\ell$ could used to represent $2^\ell$ different objects, we could store the integer from $[1,N]$ with bitstring length $\ell= \lceil \log_2N\rceil$. 
Therefore, we only need to pay a small price (note that $\log \pi(M)\in\mathcal O(\log M)$) to have a wider choices of neural network approximants. 
A formal discussion will be provided during the proof of fundamental bound theorem in the next chapter. 

Now we define the following:

\begin{definition}\label{bestMedge}
Given $d\in\mathbb N, \Omega\subset \mathbb R^d$, a function class $\mathcal C\subset L^2(\Omega)$, and an activation function $\rho:\mathbb R\to \mathbb R$, consider $\gamma>0$ where there exist $L\in\mathbb N$ and a polynomial $\pi$ such that 
$$\sup_{f\in\mathcal C}\inf_{\Phi_M \in \mathcal{NN}_{L,M,d,\rho}^\pi} \lVert f-\Phi_M\rVert_{L^2(\Omega)} \in\mathcal O(M^{-\gamma}), \quad M\to \infty.$$
The supremum of all $\gamma>0$ such that the above is satisfied by some $L$ and $\pi$ will be denoted as $\gamma_{\mathcal {NN}}^{*, e}(\mathcal C, \rho)$, referred to as the \emph{effective best $M$-edge approximation rate} by neural networks. 
\end{definition}

In the next chapter, we will prove that $\sup_{\rho}\gamma_{\mathcal {NN}}^{*,e}(\mathcal C, \rho)$ is bounded above by a quantity that only depends on $\mathcal C$. Together with the last remark below Definition \ref{bestMterm}, these two results are called the fundamental bound theorem. 

\section{Fundamental Bounds of Effective Approximation Rates}

After imposing restrictions in Definitions \ref{bestMterm} and \ref{bestMedge}, we can expect $\gamma^{*,e}(\mathcal C, \mathcal D)$ and $\gamma_{\mathcal {NN}}^{*,e}(\mathcal C,\rho)$ are no longer $\infty$. 
In fact, there is a universal quantity depending only on the compact function class $\mathcal C$ that forms the fundamental bound of the other two approximation rates. 
In words, the fundamental bound means that there is a method of approximation which only depends on the function class $\mathcal C$ itself, which can be done as good as approximation by both neural networks and by representation systems \cite{main}. 
We will state the theorems here for the sake of completeness, but only prove the ones related to neural networks.  

Now we introduce the min-max rate distortion theory and use it to define the previously described fundamental bound of a function class. 

Let $d$ be a positive integer and $\Omega$ a subset of $\mathbb R^d$, and consider function class $\mathcal C\subset L^2(\Omega)$. 
The discussion revolves the encode-decode process related to a bitstring, where a bitstring is a string of finite length consists only of digits of 0 and 1. 
Fix $\varepsilon>0$. 
We choose a positive integer $\ell$ such that every $f\in\mathcal C$ can be encoded using a bitstring of length $\ell$, in a way that after decoding the bitstring, the recovered function $\widetilde f$ satisfies $\|f-\widetilde f\|_{L^2(\Omega)}\leq\varepsilon$. 
We emphasize that the length $\ell$ of bitstring should be independent from the choices of $f$. 
To save computing resources, we are interested in the minimum of the bitstring length $\ell$ so that every function can be encoded into a bitstring as short as possible, while keeping the distortion error to be $\|f-\widetilde f\|_{L^2(\Omega)}\leq \varepsilon$.

Given a positive integer $\ell$, we define sets of binary encoders and sets of binary decoder to be
$$\mathfrak E^\ell:=\left\{E:\mathcal C\to \{0,1\}^\ell\right\}, \quad \mathfrak D^\ell:=\left\{D:\{0,1\}^\ell\to L^2(\Omega)\right\}.$$
An encoder-decoder pair $(E,D)\in \mathfrak E^\ell\times\mathfrak D^\ell$ is said to achieve uniform error $\varepsilon$ over the function class $\mathcal C$ if 
$$\sup_{f\in\mathcal C} \lVert D(E(f))-f\rVert_{L^2(\Omega)}\leq \varepsilon.$$
These motivate the definition of optimal exponent of the function class - the fundamental bound.

\begin{definition}\label{optimalrate}
Let $d\in\mathbb N, \Omega\subset \mathbb R^d,$ and $\mathcal C\subset L^2(\Omega)$. Then, for $\varepsilon>0$, the \emph{minimax code length} $L(\varepsilon,\mathcal C)$ is 
$$L(\varepsilon,\mathcal C):=\min\left\{\ell\in \mathbb N: \exists (E,D)\in\mathfrak E^\ell \times \mathfrak D^\ell: \sup_{f\in\mathcal C}\lVert D(E(f))-f\rVert_{L^2(\Omega)}\leq \varepsilon\right\}.$$
Moreover, the \emph{optimal exponent} $\gamma^*(\mathcal C)$ is defined as 
$$\gamma^*(\mathcal C):=\sup \left\{\gamma\in\mathbb R: L(\varepsilon,\mathcal C)\in\mathcal O\left(\varepsilon^{-\frac1\gamma}\right), \varepsilon\to 0\right\}.$$
\end{definition}

If $\gamma^*(\mathcal C)$ is large, then $L(\varepsilon,\mathcal C)$ will be small, hence better. If $\gamma^*(\mathcal C)=+\infty$ then $L(\varepsilon,\mathcal C)$ is bounded above as $\varepsilon\to 0$, which means a fixed minimax code length is enough for arbitrary $\varepsilon$.

As noted in the remark below Definition \ref{bestMterm}, it can be shown that $\gamma^{*,e}(\mathcal C, \mathcal D)\leq \gamma^*(\mathcal C)$ for every representation system $\mathcal D$ contained in $L^2(\Omega)$. Indeed, this is a result due to \cite{donoho} and Lemma 5.26 in  \cite{grohs}:

\begin{theorem}\label{effboundforD}
Let $d\in\mathbb N, \Omega\subset \mathbb R^d,$ and $\mathcal C\subset L^2(\Omega)$, and assume that the effective best $M$-term approximation rate of $\mathcal C$ in $\mathcal D\subset L^2(\Omega)$ is $\gamma^{*,e}(\mathcal C,\mathcal D)$. Then, we have
$$\gamma^{*,e}(\mathcal C,\mathcal D)\leq \gamma^*(\mathcal C)$$
for every representation system $\mathcal D\subset L^2(\Omega)$. 
\end{theorem}

A similar result holds for the approximation rate $\gamma_{\mathcal{NN}}^{*,e}(\mathcal C, \rho)$. We illustrate some of the proofs here, which constitutes of three intermediate lemmas. 

\begin{theorem}\label{weightnotbounded}\cite{main}
Let $d\in\mathbb N, \Omega\subset \mathbb R^d, \rho:\mathbb R\to \mathbb R, c>0,$ and $\mathcal C\subset L^2(\Omega)$. Further, let 
$$\text{\textbf{Learn}}:\left(0,\frac12\right)\times \mathcal C\to \mathcal{NN}_{\infty,\infty,d,\rho}$$
be a map such that for each pair $(\varepsilon,f)\in (0,1/2)\times \mathcal C$, every weight of the neural network $\text{\textbf{Learn}}(\varepsilon,f)$ is represented by a bitstring with length no more than $\lceil c\log_2(\varepsilon^{-1})\rceil$, while guaranteeing that 
$$\sup_{f\in \mathcal C}\lVert f-\text{\textbf{Learn}}(\varepsilon,f)\rVert_{L^2(\Omega)}\leq \varepsilon.$$
Then 
$$\sup_{f\in \mathcal C}\mathcal M(\text{\textbf{Learn}}(\varepsilon,f))\notin \mathcal O\left(\varepsilon^{-\frac1\gamma}\right), \varepsilon\to 0, \quad\text{ for all }\gamma>\gamma^*(\mathcal C).$$
\end{theorem}

Note that we demand discrete weight in neural networks. Usually the weights are not necessarily natural numbers, but it must be discrete, e.g., elements from $(1/3^{100})\mathbb Z$, or of the form $\{x_1\cdot 10^{x_2}: (x_1,x_2)\in \mathbb Z\times \mathbb Z\}$. 
In further discussion we only need the first type (equally spaced), as it is enough to benefit the approximation. 
A proof by contradiction for this theorem can be found in Proposition 3.6 of \cite{main}. 

A simple way to explain this theorem, is to note that if $\gamma$ is a real number larger than the optimal exponent of $\mathcal C$, then we will need more than $\mathcal O(\varepsilon^{-1/\gamma})$ connectivity in order for a neural network to approximate a function with error $\leq \varepsilon$. 
It is worth to note that if $M>C\varepsilon^{-1/\gamma}$, $M$ denotes the connectivity of neural networks and $C$ is a constant, then $\varepsilon>C^\gamma M^{-\gamma}$. 
Thus we can catch a glimpse of the reason why $\|f-\Phi_M\|_{L^2(\Omega)}$ is not of order $\mathcal O(M^{-\gamma})$ in Definition \ref{bestMedge}. 
This idea, although not rigorous now, serves as an important point for understanding the fundamental bound theorem. 

Observe that Theorem \ref{weightnotbounded} only works for neural network with discretized weights. To extend its use for a more general type of neural networks, we have to establish some kind of approximation by neural networks of discrete weights.  
Given a neural network with weights distributed continuously but bounded above.
We could choose a small number $\delta>0$, then choose another neural network with the same structure but tweaking each weight little bit, with at most a distance of $\delta$, so that its weights are discretized. 
We found that the settings of activation function is crucial if we want the two neural networks having approximately the same outputs.

\begin{definition}
An activation function $\rho:\mathbb R\to \mathbb R$ is \emph{acceptable} if 
\begin{enumerate}
    \item it is Lipschitz-continuous, or
    \item dominated by a polynomial, which means $\rho$ is differentiable with the existence of a polynomial $q(x)$ such that $|\rho'(x)|\leq |q(x)|$ on $\mathbb R^n$. 
\end{enumerate}
\end{definition}

The above two settings are both satisfied by the ReLU function $\rho(x)=\max\{0,x\}$ and the sigmoid function $\rho(x)=(1+e^{-x})^{-1}$, to name but a few. 
These settings are imposed so that two neural networks with approximately same weights can have a small error. 
The proof below illustrate this idea.

\begin{theorem}\label{discrete}
Let $d,L,k,M\in\mathbb N$, $\eta\in(0,1/2)$, $\Omega\subset \mathbb R^d$ be bounded, and let $\rho:\mathbb R\to \mathbb R$ be an acceptable activation function. Then, there exists a positive integer $m:= m(k,L,\rho)$ such that if $\Phi\in\mathcal{NN}_{L,M,d,\rho}$ is a neural network with connectivity and all its weights bounded (in absolute value) by $\eta^{-k}$, then there is $\widetilde{\Phi}\in\mathcal{NN}_{L,M,d,\rho}$ such that 
$$\left\lVert \widetilde{\Phi}-\Phi\right\rVert_{L^\infty(\Omega)}\leq \eta$$
and all weights of $\widetilde{\Phi}$ are elements of $\eta^m \mathbb Z\cap [-\eta^{-k}, \eta^{-k}]$. 

\end{theorem}
The proof using Lipschitz-continuous function $\rho$ has been given in \cite[p.~21]{main}. Thus, we provide a proof assuming $\rho$ is dominated by a polynomial $\pi(x)$ on $\mathbb R$ for the sake of completeness.  

\begin{proof}
Without loss of generality, we will assume the number of nonzero node weights is upper-bounded by the number of nonzero edge weights, otherwise there will be a node that does not connected by a nonzero edge weight to the next layer, hence we can replace the node weight by zero, resulted in a same neural network. 

Let $m$ be a positive integer to be specified later, we choose $\widetilde{\Phi}$ be a neural network having the same structure as $\Phi$, but with each weight replaced by a nearest element in $\eta^m\mathbb Z\cap [-\eta^{-k}, \eta^{-k}]$, so the change in each weight is no more than $\eta^m/2$. 
Moreover, for $x\in \eta^m\mathbb Z\cap [-\eta^{-k},\eta^{-k}]$, it can be expressed as $x=\eta^mn$, $n\in\mathbb Z$ with $|n|\leq \eta^{-m-k}$, hence each weight in $\widetilde{\Phi}$ can be represented using no more than $(k+m)\log_2\eta^{-1}+1$ bits (the extra bit is for the sign). 
Let $C_{max}=\eta^{-k}$. 
Let $C_W$ be the maximum of 1 and the total number of nonzero edge weights and nonzero node weights of $\Phi$, which satisfies $C_W\leq 2M\leq 2\eta^{-k}$. 
For $\ell=1,\dots,L-1$ define $\Phi^\ell:\Omega\to \mathbb R^{N_\ell}$ as 
$$\Phi^\ell(x):=\rho(W_\ell\rho(\dots\rho(W_1(x))))\quad\text{ for }x\in\Omega,$$
and $\widetilde{\Phi}^\ell$ accordingly. 
Note that $\Phi$ is not equal to $\Phi^L$ if similarly defined, as the last layer will not pass through the activation function. 
For $\ell=1,\dots,L-1$ we also let
$$e_\ell = \left\lVert\Phi^\ell-\widetilde{\Phi}^\ell\right\rVert_{L^\infty(\Omega, \mathbb R^{N_\ell})}, \quad e_L=\left\lVert\Phi-\widetilde{\Phi}\right\rVert_{L^\infty(\Omega)}.$$
Since $\Omega$ is bounded, we let $C_0$ be the maximum of 1 and $\sup\{|x|; x\in\Omega\}$. Let 
$$C_\ell=\max\left\{1,\left\lVert\Phi^\ell\right\rVert_{L^\infty(\Omega,\mathbb R^{N_\ell})}, \left\lVert\widetilde{\Phi}^\ell\right\rVert_{L^\infty(\Omega,\mathbb R^{N_\ell})}\right\}\quad\text{ for }\ell=1,\dots,L-1.$$
Now we use $W_\ell, \widetilde{W_\ell}$ to denote the affine operators on $\Phi^\ell, \widetilde{\Phi}^\ell$, respectively. Then since $\Omega$ is bounded, it is contained in some compact ball $B\subset \mathbb R^d$ of radius $r$ centered at the origin, since $W_1$ and $\widetilde{W_1}$ are continuous, both $W_1(\Omega)$ and $\widetilde{W_1}(\Omega)$ are also contained in some compact set, hence bounded. Using the same argument inductively and notice that $\rho$ is also continuous, we see $$\Omega, W_1(\Omega), \widetilde{W_1}(\Omega), W_{\ell+1}(\Phi^\ell(\Omega)), \widetilde{W_{\ell+1}}(\widetilde\Phi^\ell(\Omega)),\Phi(\Omega), \widetilde{\Phi}({\Omega}) $$
are all bounded in their respective space. Thus we can choose $R>0$ depends on $\eta$ and $k$ so that all sets above are contained in the closed ball of radius $R$ in their respective space.  

Note that for $x,y\in \mathbb R$ with $|x|,|y|\leq R$, there is $c$ between $x$ and $y$ such that $|\rho(x)-\rho(y)|=|\rho'(c)(x-y)|\leq |\pi(c)||x-y|\leq AR^n|x-y|$, hence it is reasonable to define $C_\rho=\max\{1, AR^n\}$ as a substitute of related quantity of the Lipschitz-continuous activation function in the original paper. From here we will prove the estimates:
\begin{equation}\label{mainE}e_1\leq C_0C_\rho C_W\eta^m, \text{ and }e_\ell\leq C_\rho C_WC_{\ell-1}\eta^m+C_\rho C_WC_{max}e_{\ell-1}.\end{equation}

We will proceed with induction to prove (\ref{mainE}). For the estimate of $e_1$, write $W_1(x)=B_1x+b_1$ and similarly for $\widetilde{W_1}(x)$, then the number of nonzero numbers in the column matrices $b_1,\widetilde{b_1}$ is upper-bounded by $C_W$ (recall $C_W$ is the maximum of 1 and the total number of nonzero edge weights and node weights). Hence
\begin{align*}
    e_1&=\left\lVert \Phi^1-\widetilde{\Phi}^1\right\rVert_{L^\infty(\Omega,\mathbb R^{N_1})}\\
    &=\left\lVert \rho(W_1(x))-\rho(\widetilde{W_1}(x))\right\rVert_{L^\infty}\\
    &\leq \left\lVert\rho(B_1x+b_1)-\rho(\widetilde{B_1}x+\widetilde{b_1})\right\rVert_{L^\infty} \\
    &\leq \sup_{|x|\leq R}|\rho'(x)| \|(B_1-\widetilde{B_1})x+(b_1-\widetilde{b_1})\|_{L^\infty}\\
    &\leq \sup_{|x|\leq R}|\rho'(x)| \left(\frac{\eta^m}2\cdot C_WC_0+\frac{\eta^m}2\right)\\
    &\leq C_0C_WC_\rho\eta^m.
\end{align*}
This proves the first part of (\ref{mainE}). The estimate above separate the error for $B_1-\widetilde {B_1}$ and $b_1-\widetilde{b_1}$, the term $\eta^m/2$ is the maximum difference between each entry of $b_1,B_1$ and $\widetilde{b_1}, \widetilde{B_1}$, respectively.  

In the proof below we will not be separating $B_1$ and $b_1$, assuming the reader knows what is behind the calculation. Next, we have
\begin{align*}
    e_2&=\left\lVert\Phi^2-\widetilde{\Phi}^2\right\rVert_{L^\infty}\\
    &=\left\lVert\rho(W_2(\Phi^1(x))) -\rho(\widetilde{W_2}(\widetilde{\Phi}^1(x)))\right\rVert_{L^\infty}\\
    &\leq \left\lVert\rho(W_2(\Phi^1(x))) -\rho(\widetilde{W_2}(\Phi^1(x)))\right\rVert_{L^\infty}+\left\lVert\rho(\widetilde{W_2}(\Phi^1(x)))-\rho(\widetilde{W_2}(\widetilde{\Phi}^1(x)))\right\rVert_{L^\infty}\\
    &\leq C_\rho\left\{ \left(C_W\cdot\dfrac{\eta^m}2\right)\cdot C_1 + C_W\cdot \dfrac{\eta^m}2\right\}+C_\rho C_W C_{max} e_1\\
    &\leq C_\rho C_W C_1 \eta^m+C_\rho C_WC_{\max}e_1.
\end{align*}
This agrees with the formula when $\ell=2$. Assume the formula is true for some $\ell$, then 
\begin{footnotesize}
\begin{align*}
    e_{\ell+1}&=\left\lVert\Phi^{\ell+1}-\widetilde{\Phi}^{\ell+1}\right\rVert_{L^\infty}\\
    &=\left\lVert\rho(W_{\ell+1}(\Phi^\ell(x))) -\rho(\widetilde{W_{\ell+1}}(\widetilde{\Phi}^\ell(x)))\right\rVert_{L^\infty}\\
    &\leq \left\lVert\rho(W_{\ell+1}(\Phi^\ell(x))) -\rho(\widetilde{W_{\ell+1}}(\Phi^\ell(x)))\right\rVert_{L^\infty}+\left\lVert\rho(\widetilde{W_{\ell+1}}(\Phi^\ell(x))) -\rho(\widetilde{W_{\ell+1}}(\widetilde{\Phi}^\ell(x)))\right\rVert_{L^\infty}\\
    &\leq C_\rho\left\{ \left(C_W\cdot\dfrac{\eta^m}2\right)\cdot C_\ell + C_W\cdot \dfrac{\eta^m}2\right\} + C_\rho C_W C_{max} e_\ell \\
    &\leq C_\rho C_W C_\ell \eta^m +C_\rho C_W C_{max} e_\ell.
\end{align*}
\end{footnotesize}

Until now, (\ref{mainE}) has been proved by induction. 

We also have
\begin{align*}
    e_L&=\left\lVert\Phi^L-\widetilde{\Phi}^L\right\rVert_{L^\infty}\\
    &=\left\lVert W_L(\Phi^{L-1}(x))-\widetilde{W_L}(\widetilde{\Phi}^{L-1})\right\rVert\\
    &\leq \left\lVert W_L(\Phi^{L-1}(x))-\widetilde{W_L}(\Phi^{L-1})\right\rVert_{L^\infty}+\left\lVert \widetilde{W_L}(\Phi^{L-1}(x))-\widetilde{W_L}(\widetilde{\Phi}^{L-1})\right\rVert_{L^\infty}\\
    &\leq C_W\dfrac{\eta^m}2 C_{L-1}+C_W\dfrac{\eta^m}2+C_{max}C_We_{L-1}\\
    &\leq C_WC_{L-1}\eta^m +C_WC_{max}e_{L-1}.
\end{align*}
For $\ell=1,\dots,L-1$, $0$ below is the zero vector in $\mathbb R^{N_\ell}$:
\begin{align*}
    \left\lVert\Phi^\ell(x)\right\rVert_{L^\infty(\Omega,\mathbb R^{N_\ell})}&\leq \left\lVert\Phi^\ell(x)-\rho(0)\right\rVert_{L^\infty}+\left\lVert\rho(0)\right\rVert_{L^\infty}\\
    &=|\rho(0)|+\left\lVert\rho(W_\ell\rho(\dots \rho (W_1(x))))-\rho(0)\right\rVert\\
    &\leq |\rho(0)|+C_\rho C_WC_{max} C_{\ell-1},
\end{align*}
similar result goes for $\left\lVert\widetilde{\Phi}^\ell(x)\right\rVert_{L^\infty(\Omega, \mathbb R^{N_\ell})}$, thus
$$C_\ell\leq |\rho(0)|+C_\rho C_W C_{max} C_{\ell-1}.$$
Starting from $C_1\leq |\rho(0)|+C_\rho C_W C_{max}C_0$, we can derive 
$$C_\ell\leq |\rho(0)|\sum_{k=0}^{\ell-1} (C\rho C_W C_{max})^k + C_0(C_\rho C_W C_{max})^\ell\quad\text{ for }\ell=1,\dots,L-1.$$
Then since $C_\rho C_W C_{max}\geq 1$ and $\rho(0)$ is finite, there is a fixed constant $C'>0$ such that 
$$C_\ell\leq C'C_0(C_\rho C_W C_{max})^\ell\quad\text{ for }\ell=1,\dots,L-1.$$
Now since $C_W\leq 2W\leq \eta^{-k-1}$ and $C_{max}\leq \eta^{-k}$, we can choose a fix $p\in \mathbb N$ such that 
$$C_\ell \leq C'C_0 C_\rho^\ell \eta^{-\ell (2k+1)}\leq \eta^{-p}\quad\text{ for }\ell=1,\dots,L-1.$$
To make things clear, we then choose $n\in\mathbb N$ such that 
$$\max\{C_0C_\rho C_W, C_WC_{max},C_WC_{L-1}, C_\rho C_W C_{\ell-1}, C_\rho C_W C_{max}\}\leq \dfrac{\eta^{-n}}2$$
Using previous estimates we deduce 
$$e_\ell \leq \dfrac{\eta^{-n}}2(\eta^m+e_{\ell-1})\quad\text{ for }\ell=1,\dots,L-1$$
with a convention that $e_0=0$. Furthermore, we use induction to prove there is a $r\in\mathbb N$ such that $e_\ell\leq \eta^{m-(\ell-1)n-r}$ for $\ell=1,\dots,L-1$ (in fact taking $r=n$ will suffice). First we have $e_1\leq \frac12 \eta^{m-n}\leq \eta^{m-r}$. Assume $e_\ell\leq \eta^{m-(\ell-1)n-r}$ is true for some $\ell$, then 
$$e_{\ell+1}\leq \dfrac{\eta^{-n}}2(\eta^{m}+\eta^{m-(\ell-1)n-r})\leq \frac12 (\eta^{m-n}+\eta^{m-\ell n-r})\leq \eta^{m-\ell n-r}.$$

Using the previous result, we arrive at 
$$e_L\leq \dfrac12(\eta^{m-n}+\eta^{m-(L-1)n-r})\leq \eta^{m-(L-1)n-r},$$
thus taking $m=(L-1)n+r+1=Ln+1$ will be enough to show that $e_L = \lVert \Phi-\widetilde{\Phi}\rVert_{L^\infty}\leq \eta$.
\end{proof}

The theorem above can be easily extended to the $L^2$ norm. In the end of the proof we can choose $m$ slightly larger so that $e_L=\| \Phi-\widetilde\Phi\|_{L^\infty}\leq \eta/\sqrt{C_0}$, then 
\begin{align*}
    \|\Phi-\widetilde\Phi\|_{L^2}&=\left(\int_{\Omega}(\Phi-\widetilde{\Phi})^2 dx\right)^{1/2}\\
    &\leq \|\Phi-\widetilde\Phi\|_{L^\infty} \left(\int_\Omega dx\right)^{1/2}\\
    &\leq \dfrac{\eta}{\sqrt{C_0}}\cdot \sqrt{C_0}\\
    &=\eta.
\end{align*}

Next, we state a converse of Theorem \ref{weightnotbounded} which states that if the encoder-decoder length $\ell$ remains bounded above by $\mathcal O(\varepsilon^{-1/\gamma})$ for some $\gamma>\gamma^*(\mathcal C)$, then the approximation error of all $\lVert f-\Phi\rVert$ will not converges to 0 for infinitely many $M$. Formally, we have the theorem below.

\begin{theorem}\label{conversenotbounded}\cite{main}
Let $d,L\in\mathbb N, \Omega\subset \mathbb R^d$ be bounded, $\pi$ a polynomial, $\mathcal C\subset L^2(\Omega)$, and $\rho:\mathbb R\to \mathbb R$ either Lipschitz-continuous or differentiable such that $\rho'$ is dominated by a polynomial. Then, for all $C>0$ and $\gamma>\gamma^*(\mathcal C)$, we have that 
$$\sup_{f\in\mathcal C}\inf_{\Phi\in\mathcal{NN}_{L,M,d,\rho}^\pi} \|f-\Phi\|_{L^2(\Omega)}\geq CM^{-\gamma}\quad\text{ for infinitely many }M\in \mathbb N. $$
\end{theorem}

Now we prove the fundamental bound theorem for neural networks. 

\begin{theorem}\label{funbound}\cite{main}
Let $d\in \mathbb N, \Omega\subset \mathbb R^d$ be bounded, and let $\mathcal C\subset L^2(\Omega)$. Then, for any acceptable activation function $\rho:\mathbb R\to \mathbb R$, we have
$$\gamma_{\mathcal {NN}}^{*,e}(\mathcal C,\rho)\leq \gamma^*(\mathcal C).$$
\end{theorem}
\begin{proof}
Assume the contrary that $\gamma_{\mathcal {NN}}^{*,e}(\mathcal C,\rho)>\gamma^*(\mathcal C)$, then if $\gamma\in(\gamma^*(\mathcal C), \gamma_{\mathcal {NN}}^{*,e}(\mathcal C,\rho))$, there exists $L\in\mathbb N$ and a polynomial $\pi$ such that
$$\sup_{f\in\mathcal C}\inf_{\Phi_M \in \mathcal{NN}_{L,M,d,\rho}^\pi} \lVert f-\Phi_M\rVert_{L^2(\Omega)} \in\mathcal O(M^{-\gamma}), \quad M\to \infty.$$
However, this result contradicts Theorem \ref{conversenotbounded}.
\end{proof}

\section{Transition  from  representation  system  to neural  networks}

Before continuing to the next section, we should make a few more definition to grasp the objective of the discussion. 
We already knew $\gamma_{\mathcal {NN}}^{*,e}(\mathcal C,\rho)\leq \gamma^*(\mathcal C)$, where the last quantity only depends on $\mathcal C$, and the previous quantity depends on the neural network as well. 
We will want to choose $\rho$ so that $\gamma_{\mathcal {NN}}^{*,e}(\mathcal C,\rho)$ to be as large as possible, but it is not possible to exceed $\gamma^*(\mathcal C)$. 
Therefore, we give a terminology when the two constants are equal:

\begin{definition}
Let $d\in\mathbb N, \Omega\subset \mathbb R^d$ be bounded, we say that the function class $\mathcal C\subset L^2(\Omega)$ is \emph{optimally representable by neural networks} with activation function $\rho:\mathbb R\to \mathbb R$ if 
$$\gamma_{\mathcal {NN}}^{*,e}(\mathcal C,\rho)= \gamma^*(\mathcal C).$$
\end{definition}

From the context of Theorem \ref{effboundforD}, we have similar terminology:

\begin{definition}
Let $d\in \mathbb N, \Omega\subset \mathbb R^d$, and assume that the effective best $M$-term approximation rate of $\mathcal C\subset L^2(\Omega)$ in $\mathcal D\subset L^2(\Omega)$ is $\gamma^{*,e}(\mathcal C,\mathcal D)$. If 
$$\gamma^{*,e}(\mathcal C,\mathcal D)=\gamma^*(\mathcal C),$$
then the function class $\mathcal C$ is said to be \emph{optimally representable by representation system} $\mathcal D$. 
 
\end{definition}

In this section, we aimed to establish similar connection between representation systems and neural networks. 
We treat the neural networks as a new subject, then transfer useful properties of representation systems to neural networks, in the sense that approximation by representation system are more widely studied. 

\begin{definition}\label{transition}
Let $d\in\mathbb N, \Omega\subset \mathbb R^d, \rho:\mathbb R\to \mathbb R$, and $\mathcal D=(\varphi_i)_{i\in I}\subset L^2(\Omega)$ be a representation system. Then, $\mathcal D$ is said to be \emph{representable by neural networks} (with activation function $\rho$) if there exist $L,R\in\mathbb N$ such that for all $\eta>0$ and every $i\in I$, there is a neural network $\Phi_{i,\eta}\in\mathcal {NN}_{L,R,d,\rho}$ with 
$$\|\varphi_i-\Phi_{i,\eta}\|_{L^2(\Omega)}\leq \eta.$$

If in addition, the weights of $\Phi_{i,\eta}\in\mathcal{NN}_{L,R,d,\rho}$ are  bounded above by $Ai^n\eta^{-n}$ for some $A>0, n\in\mathbb N$, and if $\rho$ is acceptable, then we say that $\mathcal D$ is \emph{effectively representable by neural networks} (with activation function $\rho$). 
\end{definition}

Note that we use $R$ instead of $M$ to denote connectivity. This is to be combined with the best $M$-term approximation from $\mathcal D$. Suppose $f$ is a function from the function class $\mathcal C$ and we are using best $M$-term approximation $\sum_{i\in I_M}c_i\varphi_i$ with $\varphi_i\in\mathcal D$ to approximate $f$. 
Then since each term from $\mathcal D$ can be approximated using a neural network of connectivity $R$, we can imagine $f$ can be approximated by a composite neural network with a total count of connectivity $RM$. 
Formally, we have the theorem below. The proof can be found in Theorem 4.2 from \cite{main}.

\begin{theorem}\label{simpletransfer}
Let $d\in\mathbb N, \Omega\subset \mathbb R^d,$ and $\rho:\mathbb R\to \mathbb R$. Suppose that $\mathcal D=(\varphi_i)_{i\in I}\subset L^2(\Omega)$ is representable by neural networks. Let $f\in L^2(\Omega)$. For $M\in\mathbb N,$ let $f_M=\sum_{i\in I_M}c_i\varphi_i$, $I_M\subset I, \# I_M=M$, satisfying
$$\|f-f_M\|_{L^2(\Omega)}\leq \varepsilon,$$
where $\varepsilon\in(0,1/2)$. Then, there exist $L\in\mathbb N$ (depending on $\mathcal D$ only) and a neural network $\Phi(f,M)\in\mathcal {NN}_{L,M',d,\rho}$ with $M'\in\mathcal O(M)$ satisfying
$$\|f-\Phi(f,M)\|_{L^2(\Omega)}\leq 2\varepsilon.$$
In particular, for all function classes $\mathcal C\in L^2(\Omega)$, it holds that 
$$\gamma_{\mathcal {NN}}^*(\mathcal C,\rho)\geq \gamma^*(\mathcal C,\mathcal D).$$
\end{theorem}

We now establish similar result for effective best $M$-term/edge approximation rates. 

\begin{theorem}\cite{main}
Let $d\in\mathbb N, \Omega \subset \mathbb R^d$ be bounded, $\rho$ an acceptable activation function, and let $\mathcal C\subset L^2(\Omega)$. Suppose that the representation system $\mathcal D=(\varphi_i)_{i\in\mathbb N}\subset L^2(\Omega)$ is effectively representable by neural networks. Then, for all $\gamma<\gamma^{*,e}(\mathcal C,\mathcal D)$, there exist a polynomial $\pi$, constants $c>0, L\in\mathbb N$, and a map
$$\textbf{Learn}:\left(0,\frac12\right)\times \mathcal C\to \mathcal{NN}_{L,\infty,d,\rho}^\pi,$$ such that for every $f\in\mathcal C$ the weights in $\textbf{Learn}(\varepsilon,f)$ can be represented by no more than $\lceil c\log_2(\varepsilon^{-1})\rceil$ bits while $\|f-\textbf{Learn}(\varepsilon,f)\|_{L^2(\Omega)}\leq \varepsilon$ and $\mathcal M(\textbf{Learn}(\varepsilon,f))\in\mathcal O(\varepsilon^{-1/\gamma})$ for $\varepsilon\to 0$. 

In particular, we have $\gamma_{\mathcal{NN}}^{*,e}(\mathcal C, \rho)\geq \gamma^{*,e}(\mathcal C, \mathcal D)$. 
\end{theorem}

\begin{proof}
Fix $\gamma<\gamma^{*,e}(\mathcal C,\mathcal D)$, let $M\in\mathbb N$. By the definition of effective best $M$-term approximation rate, there is a polynomial $\pi$, constants $C,D>0$, and $I_M\subset \{1,2,\dots,\pi(M)\}$, $\#I_M=M$, with coefficients $\max_{i\in I_M}|c_i|\leq D$ such that
$$\left\|f-\sum_{i\in I_M}c_i\varphi_i\right\|_{L^2(\Omega)}\leq \dfrac{CM^{-\gamma}}2=\dfrac{\delta_M}2,$$
where we let $\delta_M=CM^{-\gamma}$. By Definition \ref{transition} of effective representability of $\mathcal D$ by neural networks, there are $L,R\in\mathbb N$ such that for each $i\in I_M$ with $\eta:=\delta_M/\max\{1,4\sum_{i\in I_M}|c_i|\}$, there is a neural network $\Phi_{i,\eta}\in\mathcal {NN}_{L,R,d,\rho}$ satisfying 
$$\| \varphi_i-\Phi_{i,\eta}\|_{L^2(\Omega)}\leq \eta$$
with the weights of $\Phi_{i,\eta}$ bounded above by $A|i\eta^{-1}|^n$ for some constants $A>0, n\in\mathbb N$. We now define $\Phi(f,M)\in \mathcal{NN}_{L,RM,d,\rho}$ such that it is the result of $\Phi_{i,\eta}, i\in I_M$ operating in parallel then combine the outputs using the coefficients $\{c_i\}_{i\in I_M}$, defined by 
$$\Phi(f,M)=\sum_{i\in I_M}c_i \Phi_{i,\eta}.$$
This proves
$$\left\|\sum_{i\in I_M}c_i\varphi_i-\Phi(f,M)\right\|_{L^2(\Omega)}\leq \dfrac{\delta_M}4$$
Now we will represent $\Phi(f,M)$ by another neural network $\widetilde{\Phi}(f,M)\in \mathcal {NN}_{L,RM,d,\rho}$ with discrete weights by using Theorem \ref{discrete}. Since weights of $\Phi(f,M)$ is bounded above by $A|i\eta^{-1}|^n$, with $i\leq \pi(M)$ and $\eta^{-1}\leq \max\{1,4\sum_{i\in I_M}|c_i|\} \delta_M^{-1}\leq \max\{M^\gamma, DM^{\gamma+1}\}$, the weights are polynomially bounded by $\delta_M^{-1}\sim M^{\gamma}$. Thus there is $\widetilde\Phi(f,M)\in \mathcal{NN}_{L,RM,d,\rho}$ with weights represented by no more than $\lceil c\log_2\delta_M^{-1}\rceil$ bits such that
$$\|\Phi(f,M)-\widetilde{\Phi}(f,M)\|_{L^2(\Omega)}\leq \dfrac{\delta_M}4.$$
By three parts of triangle inequality, we have
$$\|f-\widetilde{\Phi}(f,M)\|_{L^2(\Omega)}\leq \delta_M=CM^{-\gamma}.$$
For $\varepsilon\in(0,1/2)$ we define 
$$\textbf{Learn}(\varepsilon,f)=\widetilde\Phi(f,M_\varepsilon)\quad\text{ with }\quad M_\varepsilon=\left\lceil\left(\dfrac{C}{\varepsilon}\right)^{1/\gamma}\right\rceil$$
Now we check the map \textbf{Learn} satisfies all of the conditions. First we have $\|f-\widetilde{\Phi}(f,M)\|_{L^2(\Omega)}\leq CM^{-\gamma}\leq \varepsilon$, then all weights of \textbf{Learn} can be represented with no more than $\lceil c\log_2(\delta_{M_\varepsilon}^{-1})\rceil$ bits, then since $\delta_{M_\varepsilon}$ has the same order as $M_\varepsilon^{-\gamma}$, which has the same order as $\varepsilon$, then the weights can be represented by no more than $\lceil c'\log_2(\varepsilon^{-1})\rceil$ bits for some $c'>0$. Since each $\Phi_{i,\eta}$ has no more than $M_\varepsilon$ connectivity, we find that $\textbf{Learn}(\varepsilon,f)$ has no more than $RM_\varepsilon$ connectivity. It is important to note from Definition \ref{transition} that $R$ doesn't depends on $\varepsilon$, hence we may regard it as a constant that doesn't depends on $M_\varepsilon$. Thus $\textbf{Learn}(\varepsilon,f)$ has no more than 
$$R(C^{1/\gamma}\varepsilon^{-1/\gamma}+1)\leq 2RC^{1/\gamma}\varepsilon^{-1/\gamma}\in \mathcal O(\varepsilon^{-1/\gamma})$$
connectivity. 
\end{proof}

The theorem shows that if $\mathcal D$ is effectively representable by neural networks, then when transfering from representation system to neural networks, we can achieve effective best $M$-edge approximation rate as $\gamma_{\mathcal{NN}}^{*,e}(\mathcal C,\rho)\geq \gamma^{*,e}(\mathcal C,\mathcal D)$. In particular, if the function class $\mathcal C$ can be optimally represented by $\mathcal D$: $\gamma^{*,e}(\mathcal C,\mathcal D)=\gamma^*(\mathcal C)$, then it can also be optimally represented by neural networks with activation function $\rho$: $\gamma_{\mathcal {NN}}^{*,e}(\mathcal C,\rho)=\gamma^*(\mathcal C)$ by fundamental bounds of effective approximation rates.

\textbf{A conclusion on theoretical results}

We have seen that in suitable settings, $\gamma^{*, e}_{\mathcal{NN}}(\mathcal C,\rho)\leq \gamma^{*}(\mathcal C)$ and $\gamma^{*,e}(\mathcal C, \mathcal D)\leq\gamma^*(\mathcal C)$ which is the  result from fundamental bound of the function class $\mathcal C$. If in addition, $\mathcal D$ is effectively representable by neural networks, then we have the double inequality: 
$$\gamma^{*,e}(\mathcal C, \mathcal D)\leq \gamma_{\mathcal{NN}}^{*,e}(\mathcal C, \rho)\leq \gamma^*(\mathcal C).$$
This inequality is interesting in its own right. From a pair of well-studied function class $\mathcal C$ and representation system $\mathcal D$, if $\mathcal D$ is effectively representable by neural networks with an acceptable activation function, then using neural networks to approximate functions in $\mathcal C$ can be done as good as using representation system, but its behavior is restricted by the optimal exponent $\gamma^*(\mathcal C)$. 

\section{Application on B-spline and cartoon-like functions}

In this section we seek practicality of using neural network to approximate B-spline functions. Then, we prove the class of $\beta$ cartoon-like functions has a finite optimal exponent. 

\subsection{Choices of activation functions}

We will need to narrow down possible choices of activation functions, so we can have a better control over the behavior of the neural networks. 

\begin{definition}\label{sigmoidal}
A continuous function $\rho:\mathbb R\to \mathbb R$ is called a \emph{sigmoidal function} of order $k\in\mathbb N, k\geq 1$, if there exists $C>0$ such that

$$\lim_{x\to -\infty}\frac1{x^k}\rho(x)=0, \lim_{x\to \infty}\frac1{x^k}\rho(x)=1, \quad\text{ and }\quad |\rho(x)|\leq C(1+|x|)^k\;\text{ for }x\in\mathbb R. $$
If in addition, $\rho$ is differentiable, then it is \emph{strongly sigmoidal} of order $k$ provided there exists constants $a,b,C>0$ such that 
$$\left|\frac1{x^k}\rho(x)\right|\leq C|x|^{-a}\text{ for } x<0; \left|\frac1{x^k}\rho(x)-1\right|\leq Cx^{-a}\text{ for }x\geq 0;$$
$$|\rho(x)|\leq C(1+|x|)^k, \left|\dfrac d{dx}\rho(x)\right|\leq C|x|^b\text{ for } x\in\mathbb R. $$
\end{definition}

It is worth to note that in practice, most activation functions are sigmoidal function of order $k=0$ or $k=1$ with a similar definition. For example, the sigmoid function 
$$\rho(x) = \frac1{1+e^{-x}},$$
and the ReLU (Rectified Linear Unit) function 
$$\rho(x) = \max\{0, x\}.$$

Although it is not common to use sigmoidal function of order $2$, let alone even higher order functions, we have include their definitions for the sake of generalization.

\subsection{Approximate B-Spline}

B-splines are classic building blocks in constructing continuous functions with compact support, that is, functions vanishes outside a compact subset of $\mathbb R^N$. The simplest B-spline with order 1, $N_1$, is the characteristic function $\chi_{[0,1]}$ defined on $\mathbb R$, where 
$$N_1(x)=\chi_{[0,1]}(x)=\begin{cases}1&\text{ if }x\in [0,1]\\
0&\text{ otherwise}.\end{cases}$$
By induction, one define a higher order B-spline $N_m$, assuming $N_{m-1}$ is known, as the convolution of $N_{m-1}$ and $N_1$. Then we have
$$N_m(x):= (N_{m-1}*N_1)(x)=\int_{\mathbb R}N_{m-1}(x-t)N_1(t)dt=\int_0^1 N_{m-1}(x-t)dt.$$
Some examples of B-splines are 
$$N_2(x)=\begin{cases}x&\text{ if }x\in[0,1],\\
2-x&\text{ if }x\in[1,2],\\
0&\text{ otherwise}.\end{cases}$$
and 
$$N_3(x)=\begin{cases}\frac12x^2&\text{ if }x\in[0,1],\\
-x^2+3x-\frac32&\text{ if }x\in[1,2],\\
\frac12x^2-3x+\frac92&\text{ if }x\in[2,3],\\
0&\text{ otherwise}.\end{cases}$$

There are many properties of the B-spline. For example, $N_m$ is $(m-2)$-times continuously differentiable nonnegative function which is identically zero outside $[0,m]$. Only continuous differentiability is not so obvious. When $m=2$ we know $N_2(x)$ is continuous. Suppose $N_{m-1}$ is $(m-3)$-times continuously differentiable, then using symmetry of convolution we have
$$N_m(x)=\int_{x-1}^x N_{m-1}(t)dt\implies N_m'(x)=N_{m-1}(x)-N_{m-1}(x-1),$$
then $N_m$ is $(m-2)$-times continuously differentiable because $N_m'$ is $(m-3)$-times continuously differentiable. There are more facts, such as for each $x\in\mathbb R$, the (finite) sum below 
$$\sum_{n=-\infty}^\infty N_m(x+n)$$
is equal to 1. Moreover, on each subinterval $[k,k+1]$ where $N_m$ is not identically zero, it is a degree $m-1$ polynomial. 

The results from \cite{chui} proved that the representation system of B-splines $\{N_m\}$ is  representable by neural networks. 
We modify the proof to show that the B-splines is effectively representable by neural networks if the activation function is strongly sigmoidal. In the discussion below, we assume $\rho$ is a fixed strongly sigmoidal function of order $k$ with related constants $a,b,C>0$. 

\begin{theorem}\label{effectiveBspline}
Let $\mathcal D=\{N_i\}$ be the representation system of B-splines. 
Let $D>0$. 
Then, there is a corresponding countable collection $\{B\Phi_{m,D,\varepsilon}\}_{m\geq 1}\subset \mathcal{NN}_{L,R,d,\rho}$ of neural networks such that 
$$\|N_m-B\Phi_{m,D,\varepsilon}\|_{L^2([-D,D])}\leq\varepsilon,$$
with the weights of $B\Phi_{m,D,\varepsilon}$ bounded above by $Am^n\varepsilon^{-n}$ for some $A>0, n\in\mathbb N$ independent of $m$, and a strongly sigmoidal function $\rho$ of order $k$.
\end{theorem}

\begin{lemma}\label{firstlemma}
Define $x_+=\max\{0,x\}$. Given $L\in\mathbb N, D>0, \varepsilon>0$, let $\rho$ be a sigmoidal activation function of order $k$, there is a neural network $\Phi_{+,\varepsilon}\in \mathcal{NN}_{L+1, L+1, 1, \rho}$ such that
$$|x_+^{k^L}-\Phi_{+,\varepsilon}(x)|\leq \varepsilon\quad\text{ for }|x|\leq D. $$
In addition, the weights of $\Phi_{+,\varepsilon}$ are bounded above by $\mathcal O(\varepsilon^{-n})$ for some positive integer $n$ that only depends on $L, k, a, b$. 
\end{lemma}
Note that the number of layers is the same as connectivity, this neural network $\Phi_{+,\varepsilon}$ only has exactly one nonzero edge weight connecting each consecutive layers, with no node weight.
\begin{proof}
Let $\delta=\left(\dfrac\varepsilon{2^{k+1}C}\right)^{1/k}$, we find $B>1$ such that $$|\rho(x)|\leq \varepsilon |x|^k\text{ for }x<-B, \quad |\rho(x)-x^k|\leq \varepsilon |x|^k\text{ for }x>B.$$
Following strong sigmoidality, we can define $B=\max\{1, (C/\varepsilon)^{1/a}\}$.
We first solve the case for $D=1, L=1$. Define $P_{1,1,\varepsilon}(x)=(\delta/B)^k \rho(Bx/\delta)\in \mathcal{NN}_{2,2,1,\rho}$, then we claim $|x_+^k-P_{1,1,\varepsilon}(x)|\leq \varepsilon$ for $|x|\leq 1$. If $x< -\delta$ then 
$$|x_+^k -P_{1,1,\varepsilon}(x)|=|P_{1,1,\varepsilon}(x)|\leq \left(\frac\delta B\right)^k \varepsilon \left(\frac{Bx}{\delta}\right)^k \leq \varepsilon|x|^k\leq \varepsilon.$$
If $-\delta\leq x<0$, then 
$$|x_+^k -P_{1,1,\varepsilon}(x)|=|P_{1,1,\varepsilon}(x)|\leq\left(\frac \delta B\right)^k \cdot C(1+B)^k\leq \dfrac\varepsilon{2^{k+1}CB^k}\cdot C(2B)^k<\varepsilon. $$
If $0\leq x\leq \delta$, then 
$$|x_+^k-P_{1,1,\varepsilon}(x)|\leq \delta^k + \left(\frac\delta B\right)^k \cdot C(1+B)^k\leq \dfrac\varepsilon2+\dfrac\varepsilon2=\varepsilon.$$
If $x> \delta$, let $y=Bx/\delta>B$ then
$$|x_+^k-P_{1,1,\varepsilon}(x)|=x^k\left|1- \left(\frac\delta{Bx}\right)^k\rho\left(\frac{Bx}\delta\right)\right|\leq |1-y^{-k}\rho(y)|\leq \varepsilon.$$

The weights of $P_{1,1,\varepsilon}$ are bounded above by the order $\mathcal O(\varepsilon^{-\frac1k-\frac1a})$. Indeed, we note that $B/\delta\in \mathcal O(\varepsilon^{-\frac1k-\frac1a})$, hence $\delta/B \in\mathcal O(1)$ by the natural assumption that $\varepsilon<1$.

For a general $D>0$, we let $$P_{1,D,\varepsilon}(x)= D^kP_{1,1,D^{-k}\varepsilon}\left(\frac xD\right)\in \mathcal{NN}_{2,2,1,\rho},$$
then for $|x|\leq D$ we have $|x/D|\leq 1$, hence 
$$|x_+^k-P_{1,D,\varepsilon}(x)|=D^k\left|\left(\frac xD\right)_+^k -P_{1,1,D^{-k}\varepsilon}\left(\frac xD\right)\right|\leq D^k(D^{-k}\varepsilon)=\varepsilon.$$

The weights of $P_{1,D,\varepsilon}$ are also bounded above by $\mathcal O(\varepsilon^{-\frac1k -\frac1a})$ because $D$ is independent from $\varepsilon$. 

The function $P_{1,1,\varepsilon}$ is continuous, hence uniformly continuous on any compact subset of $\mathbb R$. We thus choose $\eta>0$ such that 
$$|P_{1,1,\varepsilon/2}(x)-P_{1,1,\varepsilon/2}(y)|\leq \varepsilon/2\quad\text{ if }|x-y|<\eta, |x|,|y|\leq 2.$$
This is true if we choose $\eta=\frac\varepsilon{2^{b+1}C}\min\{(\frac B\delta)^{k-1-b},1\}$, where $\eta^{-1}\in \mathcal O(\varepsilon^{-N})$ for some positive integer $N\geq 1$ depends on $k,a,b$ only. Then by mean value theorem and strong sigmoidality, 
\begin{align*}
    |P_{1,1,\varepsilon/2}(x)-P_{1,1,\varepsilon/2}(y)|&=\left(\frac\delta B\right)^{k} \left|\rho\left( \dfrac{Bx}{\delta}\right)-\rho\left(\frac{By}\delta\right)\right|\\
    &\leq \left(\dfrac\delta B\right)^k \cdot C\left(\frac{2B}\delta\right)^b \cdot \dfrac B\delta |x-y|\\
    &< 2^bC\left(\dfrac \delta B\right)^{k-1-b}\eta\leq \dfrac\varepsilon2.
\end{align*}

Let $P_{2,1,\varepsilon}(x)=P_{1,1,\varepsilon/2}(P_{1,1,\eta}(x))\in \mathcal{NN}_{3,3,1,\rho}$, then for $|x|\leq 1$, 
\begin{align*}
    |x_+^{k^2}-P_{2,1,\varepsilon}(x)|&\leq |(x_+^k)^k-P_{1,1,\varepsilon/2}(x_+^k)|+|P_{1,1,\varepsilon/2}(x_+^k)-P_{1,1,\varepsilon/2}(P_{1,1,\eta}(x))|\\
    &\leq \frac\varepsilon2+\frac\varepsilon2=\varepsilon.
\end{align*}
The weights of $P_{1,1,\eta}$ are bounded above by the order $\mathcal O(\varepsilon^{-N(\frac1k+\frac1a)})$. 
Upon concatenation, we find the weights of $P_{2,1,\varepsilon}$ are bounded above by $\mathcal O(\varepsilon^{-N(\frac1k+\frac1a)})$ as well.
Suppose we already defined $P_{\ell-1,1,\varepsilon}\in \mathcal{NN}_{\ell,\ell,1,\rho}$ with weights bounded above by $\mathcal O(\varepsilon^{-N^{\ell-2}(\frac1k+\frac1a)})$, then we define $$P_{\ell,1,\varepsilon}(x)=P_{1,1,\varepsilon/2}(P_{\ell-1,1,\eta}(x))\in\mathcal{NN}_{\ell+1,\ell+1,1,\rho},$$
so that 
\begin{align*}
    |x_+^{k^\ell}-P_{\ell,1,\varepsilon}(x)|&\leq |(x_+^{k^{\ell-1}})^k-P_{1,1,\varepsilon/2}(x_+^{k^{\ell-1}})|+|P_{1,1,\varepsilon/2}(x_+^{k^{\ell-1}})-P_{1,1,\varepsilon/2}(P_{\ell-1,1,\eta}(x))|\\
    &\leq \frac\varepsilon2+\frac\varepsilon2=\varepsilon.
\end{align*}
By induction, the weights of $P_{\ell,1,\varepsilon}$ is bounded above by $\mathcal O(\varepsilon^{-N^{\ell-1} (\frac1k+\frac1a)})$.
Define $$\Phi_{+,\varepsilon}(x)=D^{k^L}P_{L,1,D^{-k^L}\varepsilon}\left( \frac xD\right)\in \mathcal{NN}_{L+1,L+1,1,\rho},$$
we have found a neural network such that $$|\Phi_{+,\varepsilon}(x)-x_+^{k^L}|\leq \varepsilon$$ for $|x|\leq D$, with weights bounded above by $\mathcal O(\varepsilon^{-N^{L-1}(\frac1k+\frac1a)})\subset \mathcal O(\varepsilon^{-n})$ for some $n$ that only depends on $L, k, a, b$. 
\end{proof}

Next, we approximate the function $x^{k^L}$. 

\begin{lemma}\label{secondlemma}
Under the assumptions of Lemma \ref{firstlemma}, there is a neural network $\Phi_{\varepsilon, D}\in \mathcal{NN}_{L+1,2L+2,1,\rho}$ such that 
$$|x^{k^L}-\Phi_{\varepsilon,D}(x)|\leq \varepsilon\quad\text{ for }|x|\leq D. $$
In addition, the weights of $\Phi_{\varepsilon,D}$ are bounded above by $\mathcal O(\varepsilon^{-n})$ for some positive integer $n$ that only depends on $L, k, a, b, D$. 
\end{lemma}
\begin{proof}
It can be verified that $x^N = x_+^N +(-1)^N(-x)_+^N$ by separating the cases whether $N$ is even or odd. 

Define 
$$\Phi_{\varepsilon,D}(x)=\Phi_{+,\varepsilon/2}(x)+(-1)^{k^L}\Phi_{+,\varepsilon/2}(-x)\in \mathcal{NN}_{L+1,2L+2,1,\rho},$$
this neural network is constructed by first feedforwarding $x$ through both $\Phi_{+,\varepsilon/2}$ in parallel, then applying one more affine function (namely, inner product) in the end, it is still $L+1$ layers because neural networks do not need to pass through any activation function in the last layer. 

Now we have
\begin{align*}
    |x^{k^L}-\Phi_{\varepsilon, D}(x)|&\leq |x_+^N-\Phi_{+,\varepsilon/2}(x)|+|(-1)^{k^L}(-x)_+^{k^L}-(-1)^{k^L}\Phi_{+,\varepsilon/2}(-x)|\\
    &\leq \frac\varepsilon2+\frac\varepsilon2=\varepsilon\quad\text{ for }|x|\leq D.
\end{align*}
Moreover, the weights of $\Phi_{\varepsilon, D}$ are still bounded above by $\mathcal O(\varepsilon^{-n})$, with the same $n$ as Lemma \ref{firstlemma}.
\end{proof}

Using techniques from linear algebra, we can approximate the function $x_+$. 
\begin{lemma}\label{thirdlemma}
Under the same assumptions in Lemma \ref{firstlemma}, given $\varepsilon>0$ there is a neural network $\Psi_{+,\varepsilon,D}(x)\in \mathcal{NN}_{2,3(k+1),1,\rho}$ such that 
$$|x_+-\Psi_{+,\varepsilon,D}(x)|\leq \varepsilon\quad\text{ for }|x|\leq D. $$

In addition, the weights of $\Psi_{+, \varepsilon,D}$ are bounded above by $\mathcal O(\varepsilon^{-n})$ for some positive integer $n$ that only depends on $k$ and $a$. 

\end{lemma}
\begin{proof}
The first step is to find constants $\alpha_0,\alpha_1,\dots,\alpha_k$ such that 
$$\sum_{\mu=0}^k \alpha_\mu (x+\mu)^k=x\quad \forall x\in\mathbb R,$$
where we emphasize $k$ is the sigmoidal order of the activation function $\rho$. By expanding coefficients, we have $(x+\mu)^k=\sum_{v=0}^k \binom kv x^v\mu^{k-v}$, hence 
\begin{align*}
    \sum_{\mu=0}^k \alpha_\mu \mu^{k-v}=\begin{cases}1/k&\text{ if }v=1,\\0&\text{ otherwise.}\end{cases}
\end{align*}
This is a linear system in $k+1$ equations with full rank because the coefficient matrix on the left is Vandermonde. Therefore, the solution $\{\alpha_0,\alpha_1,\dots,\alpha_k\}$ exists. Observe the scalars $\{\alpha_0,\dots,\alpha_k\}$ only depends on $k$. 

Define $N$ to be the smallest positive integer such that $$N\geq \max\left\{\dfrac{2k^k\sum_\mu |\alpha_\mu|}\varepsilon, k\right\},$$
Then for $x>0$ we have
$$\sum_{\mu=0}^k \alpha_\mu N^{k-1} \left(x+\frac\mu N\right)_+^k=\frac1N \sum_{\mu=0}^k \alpha_\mu(Nx+\mu)^k=\frac1N\cdot Nx=x=x_+.$$
Observe that the above also holds for $x\leq -k/N$ because both sides are zero. 

If $-k/N<x\leq 0$ then $\dfrac{\mu-k}N<x+\dfrac\mu N\leq \dfrac\mu N$ and $(x+\mu/N)_+\leq |x+\mu/N|\leq k/N$. We have
$$\left|\sum_{\mu=0}^k \alpha_\mu N^{k-1} \left(x+\frac\mu N\right)_+^k\right|\leq \left(\frac kN\right)^k N^{k-1}\sum_{\mu=0}^k |\alpha_\mu|\leq \frac\varepsilon2.$$
Thus, it is true that 
$$\left|\sum_{\mu=0}^k \alpha_\mu N^{k-1} \left(x+\frac\mu N\right)_+^k-x_+\right|\leq \frac\varepsilon2\quad\forall x\in \mathbb R.$$
We then let $\eta=\varepsilon/(2N^{k-1}\sum |\alpha_\mu|)$ and define 
$$\Psi_{+,\varepsilon,D}(x)=\sum_{\mu=0}^k N^{k-1}\alpha_\mu P_{1,D+1,\eta}(x+\mu/N)\in \mathcal{NN}_{2,3(k+1),1,\rho}.$$
Two points need to be clarified. Why do we choose $P_{1,D+1,\eta}$ instead of $P_{1,D,\eta}$? It is because we have $N\geq k$ so $0\leq \mu/N\leq 1$, hence the domain for $x+\mu/N$ should be $(-D-1,D+1)$ to account for the extra length of $\mu/N$. Moreover, we know $\Psi_{+,\varepsilon,D}$ is equivalent to $k+1$ neural networks in $\mathcal{NN}_{2,2,1,\rho}$ operating in parallel then taking linear combinations at the end, but why $\Psi_{+,\varepsilon}$ is not necessarily a member of $\mathcal{NN}_{2,2(k+1),1,\rho}$? We note that each subnetwork $P_{1,D+1,\eta}$ does not just take an input $x$, but $x+\mu/N$. The extra $\mu/N$ should be treated as a nonzero node weight (when $\mu>0$) at the first hidden level, hence contribute one more count for the number of weights. 

We see 
\begin{align*}
    &\left|\Psi_{+,\varepsilon,D}(x)-\sum_{\mu=0}^k \alpha_\mu N^{k-1}\left(x+\frac\mu N\right)_+^k\right|\\
    \leq& \sum_{\mu=0}^k |\alpha_\mu|N^{k-1}\left|P_{1,D+1,\eta}\left(x+\frac\mu N\right)-\left(x+\frac\mu N\right)_+^k\right|\\
    \leq& \eta\sum_{\mu=0}^k |\alpha_\mu|N^{k-1}\\
    =&\dfrac\varepsilon2\quad\text{ for }|x|\leq D.
\end{align*}
We conclude $|\Psi_{+,\varepsilon,D}(x)-x_+|\leq \varepsilon$ for $|x|\leq D$. 
Moreover, we observe $\eta^{-1}\in \mathcal O(\varepsilon^{-1}\cdot \varepsilon^{-(k-1)})=\mathcal O(\varepsilon^{-k})$ since $N\in\mathcal O(\varepsilon^{-1})$, hence the weights of $\Psi_{+,\varepsilon,D}$ are dominated by $\mathcal O(\max\{\varepsilon^{-k(\frac1k+\frac1a)}, \varepsilon^{-(k-1)}\})$. 
\end{proof}

Now we summarize the approximants with their corresponding monomials so far. 
$$\Phi_{\varepsilon,D}\text{ used to approximate }x^{k^L}\text{ with error }\varepsilon,\text{ for }|x|\leq D.$$
$$\Psi_{+,\varepsilon,D}\text{ used to approximate }x_+\text{ with error }\varepsilon,\text{ for }|x|\leq D.$$

In the next lemma we use neural network to approximate monomials of any positive integer degree. 
\begin{lemma}\label{fourthlemma}
For $D,\varepsilon>0, m\in\mathbb N$, $\rho$ a sigmoidal function of order $k$. Let $L$ be the smallest integer such that $m-1\leq k^L$. Let $M=(k^L+1)(3k+2L+8)$. We construct a network $R_{D,m,\varepsilon}\in \mathcal{NN}_{L+2,M,1,\rho}$ such that 
$$|x_+^{m-1}-R_{D,m,\varepsilon}(x)|\leq \varepsilon\quad\text{ for }|x|\leq D.$$
In addition, the weights of $R_{D,m,\varepsilon}$ are bounded above by $\mathcal O(m^N\varepsilon^{-N})$ for some positive integer $N$ that only depends on $L,k,a,b,D$. 
\end{lemma}
\begin{proof}
Using the same idea as in Lemma \ref{thirdlemma}, we solve $\{a_i\}_{0\leq i\leq k^L}$ from
\begin{align*}
    \sum_{i=0}^{k^L} a_i i^{k^L-v}=\begin{cases}\dbinom{k^L}{m-1}^{-1}&\text{ if }v=m-1,\\ 0&\text{otherwise.}\end{cases}
\end{align*}
The solutions $\{a_i\}$ do not depend on $\varepsilon$ and $D$, but it depends on $m$.
In fact, we write the system above as $Va=b$, where $V$ is the $(k^L+1)\times (k^L+1)$ Vandermonde matrix and $a=[a_0\dots a_{k^K}]^T$. Then $V$ depends on $m$ because $L$ depends on $m$, and we have $L\in \mathcal O(\log m)$. Since $b$ has all entries not larger than 1, $b$ is independent from $m$. 
We have $a=V^{-1}b$ and we claim that $\{a_i\}$ are bounded above by $\mathcal O(m^{n'})$ for some positive integer $n'$. 
Indeed, \textcolor{blue}{\href{https://proofwiki.org/wiki/Inverse_of_Vandermonde_Matrix}{this proof}} gives a formula for inverse of a Vandermonde matrix, each entry in $V^{-1}$ is bounded above by $\mathcal O((k^L)^{n'})=\mathcal O(m^{n'})$ for some positive integer $n'$. 

Next, we see $x^{m-1}=\sum_{i=0}^{k^L} a_i(x+i)^{k^L}$ for all real $x$. 
In particular, the expression is also true if we replace $x$ by $x_+$. Let $$\eta=\dfrac{\varepsilon}{2k^L(k^L+2)^{k^L-1} \sum_i |a_i|},$$
we define 
$$R_{1,m,\varepsilon}(x)=\sum_{i=0}^{k^L} a_i \Phi_{\eta, k^L+2}(\Psi_{+,\eta,1}(x)+i)\in \mathcal{NN}_{L+2,M,1,\rho},$$
hence the weights of $R_{1,m,\varepsilon}$ is affected by $m$.
For $|x|\leq 1$, we note that if $i=0,1,\dots, k^L$, then
$$|(\Psi_{+,\eta,1}(x)+i)-(x_++i)|\leq \eta.$$
Since $|x_++i|\leq k^L+1$, for $\eta$ small enough (by choosing $\varepsilon$ small enough) we have
$$|\Psi_{+,\eta,1}(x)+i|\leq k^L+2.$$
Now note that if $a,b$ are two real numbers, by mean value theorem 
$$|a^{k^L}-b^{k^L}|\leq k^L \max\{|a|,|b|\}^{k^L-1}|a-b|,$$
thus we have
$$\left|(\Psi_{+,\eta,1}(x)+i)^{k^L}-(x_++i)^{k^L}\right|\leq k^L(k^L+2)^{k^L-1}\eta.$$
We also have 
$$|\Phi_{\eta, k^L+2}(\Psi_{+,\eta,1}(x)+i)-(\Psi_{+,\eta,1}(x)+i)^{k^L}|\leq \eta.$$
Combining the two inequalities above, we have
$$|\Phi_{\eta, k^L+2}(\Psi_{+,\eta,1}(x)+i)-(x_++i)^{k^L}|\leq 2k^L(k^L+2)^{k^L-1}\eta.$$
To conclude, we observe that
\begin{align*}
    \left|R_{1,m,\varepsilon}(x)-\sum_{i=0}^{k^L}(x_++i)^{k^L}\right|&\leq 2k^L(k^L+2)^{k^L-1}\eta\sum_i |a_i|\leq \varepsilon\quad\text{ for }|x|\leq 1.
\end{align*}
For a general $D>0$, we define
$$R_{D,m,\varepsilon}(x)=D^{m-1}R_{1,m,D^{-m+1}\varepsilon}\left(\frac xD\right),$$
then it follows that 
$$|x_+^{m-1}-R_{D,m,\varepsilon}(x)|\leq \varepsilon\quad\text{ for }|x|\leq D.$$
Moreover, the weights of $R_{D,m,\varepsilon}$ are bounded above by $\mathcal O(m^n\eta^{-n})$ for some positive integer $n$ depends on $L,k,a,b,D$, which in turn bounded above by $\mathcal O(m^N\varepsilon^{-N})$ for another positive integer $N$ depends on $L,k,a,b,D$ because $\varepsilon\in\mathcal O(\eta m^{n'})$.
\end{proof}

Now we show $N_m(x)$ can be approximated by a neural network with error $\varepsilon$, while its weights are bounded above by $\mathcal O(m^n\varepsilon^{-n})$. 
Thus this will prove the representation system $\{N_m\}$ can be effectively represented by neural networks. The theorem below is a reformulation of Theorem \ref{effectiveBspline}, stated with more details. 

\begin{theorem}\label{B-spline}
For $L, m, k\in \mathbb N$ with $m\geq 2$, $\varepsilon, D>0$, define $M=(m+1)(k^L+1)(3k+2L+8)$, $\rho$ a sigmoidal function of order $k$. Then there is a network $B\Phi_{m,D,\varepsilon}\in\mathcal{NN}_{L+2, M, 1, \rho}$ such that 
$$\|N_m-B\Phi_{m, D,\varepsilon}\|_{L^2([-D,D])}\leq \varepsilon.$$
In addition, the weights of $B\Phi_{m,D,\varepsilon}$ are bounded above by $\mathcal O(m^n\varepsilon^{-n})$ for some positive integer $n$ that only depends on $L,k,a,b,D$. 
\end{theorem}
\begin{proof}
The first step is to present $N_m$ in a different way. We claim that 
$$N_m(x)=\frac1{(m-1)!}\sum_{j=0}^m \binom mj (-1)^j (x-j)_+^{m-1}\quad\text{ for }m\geq 2.$$
The B-spline $N_2(x)$ is the hat function, defined by $x$ when $x\in[0,1]$ and $2-x$ when $x\in[1,2]$. By carefully consider each unit interval we see
$$N_2(x)=x_+-2(x-1)_++(x-2)_+.$$

To prove for $m>2$ we use induction. Suppose the summation is true for $m=n-1$, then using convolution we have
\begin{align*}
    N_n(x)&=\int_0^1 N_{n-1}(x-t)dt\\
    &=\dfrac1{(n-2)!}\sum_{j=0}^{n-1}\binom{n-1}j (-1)^j \int_0^1 (x-j-t)_+^{n-2}dt\\
    &=\frac1{(n-1)!}\sum_{j=0}^{n-1}\binom{n-1}j (-1)^j[(x-j-t)_+^{n-1}]_1^0\\
    &=\frac1{(n-1)!}\sum_{j=0}^{n-1}\binom{n-1}j (-1)^j[(x-j)_+^{n-1}-(x-j-1)_+^{n-1}]\\
    &=\frac1{(n-1)!}\Bigg\{x_+^{n-1}+\sum_{j=1}^{n-1}\left[\binom{n-1}j+\binom{n-1}{j-1}\right](-1)^j (x-j)_+^{n-1}\\
    &\qquad +\binom{n-1}{n-1}(-1)^n(x-n)_+^{n-1}\Bigg\}\\
    &=\frac1{(n-1)!}\sum_{j=0}^n \binom nj(-1)^j (x-j)_+^{n-1}
\end{align*}
For simplicity we write $N_m(x)=\sum_{j=0}^m b_j (x-j)_+^{m-1}$ with $b_j=\frac1{(m-1)!}\binom mj(-1)^j$, then we also have $\sum|b_j|=2^m/(m-1)!$. 

Define $\eta=\varepsilon(\sqrt{2D}\sum|b_j|)^{-1}=\varepsilon\dfrac{(m-1)!}{2^{m}\sqrt{2D}}$, using Lemma \ref{fourthlemma} we define
$$B\Phi_{m,D,\varepsilon}(x)=\sum_{j=0}^m b_j R_{D+m,m,\eta}(x),$$
then for $|x|\leq D$, 
\begin{align*}
    |N_m(x)-B\Phi_{m, D, \varepsilon}|&=\left|\sum_{j=0}^m b_j [(x-j)_+^{m-1}-R_{D+m, m, \eta}(x)]\right|\\
    &\leq \eta\sum_{j=0}^m |b_j|\\
    &=\dfrac{\varepsilon}{\sqrt{2D}}. 
\end{align*}
Then, the $L^2$ norm is estimated as
\begin{align*}
    \|N_m-B\Phi_{m,D,\varepsilon}\|_{L^2([-D,D])}^2&=\int_{-D}^D (N_m(x)-B\Phi_{m,D,\varepsilon}(x))^2dx\leq \varepsilon^2,
\end{align*}
which proves the $L^2$ error is bounded above by $\varepsilon$.
Finally, by the previous lemma, the weights of $B\Phi_{m,D,\varepsilon}$ is bounded above by $\mathcal O(m^n \eta^{-n})\subset O(m^n \varepsilon^{-n})$ for some positive integer $n$ because $\eta^{-1}\leq \varepsilon^{-1}$ for sufficiently large $m$. 
\end{proof}

As a final remark, since $\mathcal D=\{N_m\}$ is effective representable by neural networks, if we let $\mathcal C$ be the class of B-spline curves, then 
$$\gamma^{*,e}(\mathcal C, \mathcal D) \leq \gamma_{\mathcal{NN}}^{*,e}(\mathcal C, \rho)\leq \gamma^*(\mathcal C).$$
This sums up the goodness of approximation of B-spline curves by neural networks. In the next section we focus on the class $\mathcal C$ of beta cartoon-like functions, calculating its optimal exponent $\gamma^*(\mathcal C)$. 

\section{Approximate cartoon-like function}

The result of this section is mainly due to \cite{donoho2}. 
We show a function class containing the \quotes{$\beta$ cartoon-like functions} has finite optimal exponent equals to $\beta/2$, then we illustrate the proofs.
Recall from Definition \ref{optimalrate}, we know optimal exponent is a quantity intrinsic to the function class.
Although this result is not about neural networks or about representation system, we should remark that by fundamental bound theorem, both effective best $M$-edge/$M$-term approximation rate for the class of $\beta$ cartoon-like functions are bounded above by $\beta/2$. 

One major application of neural networks is its ability to approximate cartoon-like function. 
Roughly speaking, we consider a function $f:[0,1]^2\to \mathbb R$ which is twice continuously differentiable except on some simple closed curve in $I:=[0,1]^2$, satisfying some regularity conditions. 
We consider this type of functions for a natural reason.
Most image pixels are not randomly placed, they are connected in a nice way so that the image has some smooth regions and some boundaries. In this section we only discuss about square containing a single simple closed curve, as this result could be generalized to the cases where multiple curves appear in a square.

First we define the curves in $I$. Generally, we pick an interior point $b_0\in I$ to serve as the center of the closed curve, then use polar coordinate to define the curve. 
For $\theta\in[0,2\pi)$ define $\rho(\theta)>0$ be a radius aparts from $b_0$, and that $\rho(0)=\lim_{\theta\to 2\pi-} \rho(\theta)$ (periodic). 
In Cartesian coordinate, the curve is given by
$$\alpha(\theta)=b_0+(\rho(\theta)\cos\theta, \rho(\theta)\sin\theta)\quad \forall  \theta\in[0,2\pi).$$
Then, for $\beta\in(1,2]$, define  $\text{HÖLDER}^\beta(C)$ to be the collection of continuously differentiable polar radius functions $\rho(\theta)$ satisfying the \textbf{boundary regularity condition} below: 
$$|\rho'(\theta_1)-\rho'(\theta_2)|\leq C |\theta_1-\theta_2|^{\beta-1}.$$

The infimum of $C>0$ such that the above holds is denoted by $\|\rho\|_{\dot\Lambda^\beta}$. Thus $\|\rho\|_{\dot\Lambda^\beta}\leq C$ for every $\rho\in\text{HÖLDER}^\beta(C)$.

Denote $B(\rho)$ to be the closed region enclosed by the curve $\alpha(\theta)$. 
Specifically, we are interested in the following set of curves:
\begin{footnotesize}
$$\text{STAR-SET}^\beta(C)=\left\{B(\rho): B(\rho)\subset \left[\frac1{10},\frac9{10}\right]^2, \frac1{10}\leq \rho(\theta)\leq\frac12, \theta\in [0,2\pi), \rho\in \text{HÖLDER}^\beta(C)\right\}.$$
\end{footnotesize}

The function class we are working in will be 
$$\text{STAR}^\beta(C)=\{ \chi_{B(\rho)}: B(\rho)\in \text{STAR-SET}^\beta(C)\},$$
where $\chi_E$ is the characteristic function which is defined by 
$$\chi_E(x)=\begin{cases}1&\text{ if }x\in E,\\0&\text{ if }x\notin E.\end{cases}$$
Since the area of $B(\rho)$ is smaller than 1, we have $\|f\|_2<1$ for all $f\in\text{STAR}^\beta(C)$.

The following is our main result:
\begin{theorem}\label{betacurve}
For $C>0$ and $\beta\in (1,2]$, we have
$$\gamma^*(\text{\textup{STAR}}^\beta(C))=\dfrac\beta2.$$
\end{theorem}
The set $\text{STAR}^\beta(C)$ is a much smaller subset of $\mathcal E^\beta(\mathbb R^2, v)$ in \cite[Theorem 6.3]{main}. The result of the theorem is also different from \cite[Theorem 1]{donoho2} because the measurement of optimality is different. However, we prove that the essential results are all equivalent. Moreover, we exclude the case $\beta=1$ since we want the boundary to be continuously differentiable.

In order to prove Theorem \ref{betacurve}, we need four steps.
\begin{enumerate}[1.]
    \item An elegant fact from rate-distortion theory \cite{berger} which we are not proving here. 
    Suppose we have $m$ coins, each coin has equal probability to render head or tail when flipped. Then there are $2^m$ possibilities of flipping these $m$ coins in a row. 
    Each result from a flip of $m$ coins can be encoded into a bitstring of length $m$, consists of zeros and ones only, then we use the obvious way to recover the state of the $m$ coins. 
    We are interested in encoding the flipped coins into bitstring of length $R<m$, then design a decoding process to recover the result of flipped coins while accepting some information losses. 
    If $X$ is the original sequence of $m$ coins, we use $\hat X=\Dec(\Enc(X))$ to denote the sequence after recovery. 
    The loss of information is simply $\text{Dist}(X,\hat X)$, the number of different heads and tails before and after recovery. 
    By rate-distortion theory, when $R$ is significantly smaller than $m$, the recovery must suffer a great amount of loss. There is a positive number $D_m(R)$ such that
    $$\inf_{\Enc, \Dec}\text{Average}(\text{Dist}(X, \Dec(\Enc(X))))\geq D_m(R)$$ where the infimum is taken across all encoding-decoding process, and the average is taken on all $2^m$ possible sequences of coins flipped. 
    The rate-distortion theory guarantees that given a positive number $\rho<1/2$, there is a number $D_1(\rho)>0$ such that $R\leq \rho m$ implies $D_m(R)\geq D_1(\rho)m$. This means if we are using considerably less resource to encode the $m$ flipped coins, then a substantial fraction of information will be lost no matter how good the encoding-decoding process is (which agrees with common sense).
    
    \item In the second step, we prove $\text{STAR}^\beta(C)$ contains a copy of $p$-hypercube, denoted by $\ell_0^p$ (which will be defined later) for $p=\frac2{\beta+1}$.
    \item Next, we show if a function class $\mathcal F$ contains a copy of $\ell_0^p$ then $\gamma^*(\mathcal F)\leq \frac{2-p}{2p}$ from the result of rate-distortion theory. 
    Thus when $\mathcal F=\Star$ we deduce that $$\gamma^*(\mathcal F)\leq \dfrac{2-\frac2{\beta+1}}{\frac{4}{\beta+1}}=\frac\beta2.$$
    \item In the last step, we prove $\gamma^*(\text{STAR}^\beta(C))\geq \dfrac\beta2$ by using the discrete wedgelets dictionary. 
\end{enumerate}

We should look at Step 2 and Step 3 at the same time. 
Note that  $\gamma>\gamma^*(\Star)$ if for any $\varepsilon\in(0,1/2)$, the minimax code length $L(\varepsilon, \Star)$ does not obey the order of $\mathcal O(\varepsilon^{-1/\gamma})$ as $\varepsilon\to 0$ (recall Definition \ref{optimalrate}). 
Our objective is to prove this is true for every $\gamma>\dfrac{2-p}{2p}$ with $p=\dfrac2{\beta+1}$, thus implies the inequality $\gamma^*(\Star)\leq \dfrac{2-p}{2p}$.
Since the statement about minimax code length is a negative statement, we do not need to use the whole class $\Star$. Instead, it is enough to rely only on the orthogonal hypercube structure embedded in $\Star$. To see why hypercube is related, we recall the description of rate-distortion theory in Step 1. It mentioned the information loss when recording a bitstring (or a sequence of coins flipped). Hypercube also has this property, as we can regard each vertice of an $m$-dimensional hypercube as a bitstring of length $m$.  

\begin{definition}
A function class $\mathcal F$ is said to contain an \emph{embedded orthogonal hypercube} of dimension $m$ and side $\delta$ if there exist $f_0\in\mathcal F$, and orthogonal functions $\psi_{i,m,\delta}$, $i=1,\dots,m$, with $\|\psi_{i,m,\delta}\|_{L^2}=\delta$, such that the collection of hypercube vertices
$$\mathcal H(m; f_0,(\psi_i)) = \left\{h=f_0+\sum_{i=1}^m \xi_i \psi_{i,m,\delta},\quad  \xi_i\in\{0,1\}\right\}$$
is embed in $\mathcal F$, i.e., $\mathcal H(m; f_0,(\psi_i))\subset \mathcal F$. 
\end{definition}
Note that orthogonality is meant by $\int \psi_{i,m,\delta}\psi_{j,m,\delta}=0$ for $i\neq j$. Moreover, it should be emphasized that $\mathcal H$ only contains the vertices of the hypercube. 

\begin{definition}\label{hypercube}
A function class $\mathcal F$ \emph{contains a copy of $\ell_0^p$} if $\mathcal F$ contains embedded orthogonal hypercubes of dimension $m(\delta)$ and side $\delta$, and if, for some sequence $\delta_k\to 0$, and some constant $C>0$: 
$$m(\delta_k)\geq C\delta_k^{-p}, \quad k=k_0,k_0+1,\dots$$
and also $\{m(\delta_k)\}_{k\geq k_0}=n_0+\mathbb N$ for some positive integer $n_0$. 
\end{definition}

This definition states clear that if the side length of the hypercube is small, then the dimension should be sufficiently large. The last condition requires the sequence $\{m(\delta_k)\}$ not to be too sparse, which was overlooked in the original paper \cite{donoho}. 

Now we prove $\Star$ contains a copy of $\ell_0^p$ for $p=2/(\beta+1)$. In short, we construct $m$-dimensional hypercubes consists of $m$ \quotes{flower petals}, since they have disjoint interior, they are orthogonal in the sense of inner product in $L^2([0,1]^2)$. 

\begin{theorem}\label{secondstep}
The function class $\text{\textup{STAR}}^\beta(C)$ contains a copy of $\ell_0^p$ for $p=2/(\beta+1)$. 
\end{theorem}
\begin{proof}
Let $\varphi\geq 0$ be a smooth function with compact support $\subset [0,2\pi]$, and also have $\varphi(0)=\varphi(2\pi)$. For example one can choose $\varphi(\theta)=\sin\dfrac\theta 2$ on $[0,2\pi]$. We will use this function as a generator to generate a copy of $\ell_0^p$ in $\text{STAR}^\beta(C)$.

We see for $a,b\in[0,2\pi]$, 
$$|\varphi'(a)-\varphi'(b)|=\frac12\left|\cos\frac a2-\cos\frac b2\right|\leq \dfrac14|a-b|\leq R|a-b|^{\beta-1}$$
for some number $R$ only depends on $\beta$, hence $\|\varphi\|_{\dot\Lambda^\beta}$ exists and is bounded above by $R$. We also have $\|\varphi\|_{L^1}=\int_0^{2\pi} \sin\frac\theta2 d\theta = 4$. For $\delta>0$ choose 
$$m=m(\delta):=\left\lfloor\left(\dfrac{\delta^2}C\dfrac{\|\varphi\|_{\dot\Lambda^\beta}}{\|\varphi\|_{L^1}}\right)^{-1/(\beta+1)}\right\rfloor,\quad A=A(\delta, C):=\dfrac{\delta^2 m^{\beta+1}}{\|\varphi\|_{L^1}}$$

as in \cite{donoho2}. We define $$\varphi_{i,m}(t) = Am^{-\beta} \varphi(mt-2\pi i)\quad\text{ for }i=1,2,\dots,m$$
so $\varphi_{i,m}$ is only supported in $[2\pi i/m, 2\pi (i+1)/m]$. This implies $\{\varphi_{i,m}\}_{i=1}^m$ is orthogonal. We also have
\begin{align*}|\varphi_{i,m}'(a)-\varphi_{i,m}'(b)|&=Am^{1-\beta}|\varphi'(ma-i)-\varphi'(mb-i)|\\
&\leq Am^{1-\beta} \|\varphi\|_{\dot\Lambda^\beta} |ma-mb|^{\beta-1}\\
&=A\|\varphi\|_{\dot\Lambda^\beta} |a-b|^{\beta-1}\end{align*}
This shows $\varphi_{i,m}$ satisfies the boundary regularity condition, and from the settings of $m$ and $A$,  $$\|\varphi_{i,m}\|_{\dot\Lambda^\beta}\leq A\|\varphi\|_{_{\dot\Lambda^\beta}}\leq C.$$ 

Fix an origin at $b_0=(1/2,1/2)$, let $r_0=1/4$ and $f_0=1_{\{r\leq r_0\}}$ be the characteristic function of the circle of radius $r_0$ centered at $b_0$. Then we define
$$\psi_{i,m}=1_{\{r\leq \varphi_{i,m}+r_0\}}-f_0, \quad i=1,2,\dots,m$$
where $1_{\{r\leq \varphi_{i,m}+r_0\}}$ is the characteristic function of the region with radius function $\varphi_{i,m}+r_0$ centered at $b_0$. 
The graph of $\psi_{i,m}$ can be treated as one of the $m$ petals of a flower centered at $b_0$. 
The collection $\{\psi_{i,m}\}_{i=1}^m$ are orthogonal in $L^2([0,1]^2)$ because $\{\varphi_{i,m}\}_{i=1}^m$ has disjoint support on $[0,2\pi]$. 
For $\psi_{i,m}$ to be an element of $\text{STAR}^\beta(C)$, its radius function should between $1/10$ and $1/2$, hence we require $\varphi_{i,m}\leq 1/4$ as well. 
We show this can be true if $\delta$ is small enough. For $t\in\mathbb R$, 
\begin{align*}
    \phi_{i,m}(t)&\leq Am^{-\beta}\\
    &=\dfrac{\delta^2 m}{\|\varphi\|_{L^1}}\\
    &\leq C' \delta^{\frac{2\beta}{\beta+1}}\leq \frac14
\end{align*}
for a small $\delta$, hence every smaller $\delta$. 

For each bitstring $\xi=(\xi_1,\dots,\xi_m)\subset \{0,1\}^m$ of length $m$, we have a correspondence between radius functions and characteristic functions: 
$$r_\xi = \dfrac14+\sum_{i=1}^m \xi_i \varphi_{i,m}\iff f_\xi =f_0+\sum_{i=1}^m \xi_i\psi_{i,m},$$
which means if the region centered at $b_0=(1/2,1/2)$ is enclosed by the radius function $r_\xi$, then its characteristic function is $f_\xi$. 

Now we know $\{\psi_{i,m}\}$ is a collection of orthogonal functions because they are supported on disjoint region, each of them has length $\Delta = \|\psi_{i,m}\|_2$, being independent of $i$ and satisfying
$$\Delta^2 = \|\psi_{i,m}\|_2^2 = Am^{-\beta} \int_{i/m}^{(i+2\pi)/m}\varphi(mt-i)dt=Am^{-\beta-1} \|\varphi\|_{L^1}=\delta^2,$$
hence $\Delta=\delta$. We have verified the functions $\{\psi_{i,m}\}$ are orthogonal, having lengths $\delta$, satisfying the boundary regularity conditions. Therefore, the hypercube $\mathcal H(m; f_0, (\psi_{i,m}))$ can be embedded in $\text{STAR}^\beta(C)$. 

Now we choose $\delta_0>0$ small enough such that $C'\delta_0^{\frac{2\beta}{\beta+1}}\leq1/4$ and also
$$\left\lfloor\left(\dfrac{\delta_0^2}C\dfrac{\|\varphi\|_{\dot\Lambda^\beta}}{\|\varphi\|_{L^1}}\right)^{-1/(\beta+1)}\right\rfloor\geq \frac12 \left(\dfrac{\delta_0^2}C\dfrac{\|\varphi\|_{\dot\Lambda^\beta}}{\|\varphi\|_{L^1}}\right)^{-1/(\beta+1)},$$
Note that the right-hand side of above increases as $\delta\to 0$. For $\delta\in(0,\delta_0)$, 
$$m(\delta)\geq \frac12 \left(\dfrac{\delta_0^2}C\dfrac{\|\varphi\|_{\dot\Lambda^\beta}}{\|\varphi\|_{L^1}}\right)^{-1/(\beta+1)}=C_0 \delta^{-\frac2{\beta+1}}$$
for some constant $C_0$ depending only on $\beta, C$ and the choice of $\varphi$. We also choose a sequence $\{\delta_k\}\subset (0,\delta_0)$ such that $m(\delta_{k+1})-m(\delta_k)=1$ for all $k\geq k_0$ sufficiently large, as it is clearly possible from the definition of $m(\delta)$. 

\end{proof}

Now we prove an upper bound for $\gamma^*(\text{STAR}^\beta(C))$. 
\begin{theorem}\label{thirdstep}
If a function class $\mathcal F$ contains a copy of $\ell_0^p$, $p<2$, and there is $A>0$ such that $\|f\|_{L^2}\leq A$ for all $f\in\mathcal F$, then $\gamma^*(\mathcal F)\leq \dfrac{2-p}{2p}$. 

We assume $\mathcal F$ has an orthonormal countable basis $\{\varphi_i\}$ which are not necessarily members of $\mathcal F$, in the sense that $\|\varphi_i\|_{L^2}=1, \int \varphi_i\varphi_j=0$ for $i\neq j$, and that every $f\in\mathcal F$ can be expressed as 
$$f(x)=\sum_{i=1}^\infty \theta_i \varphi_i(x)$$
for real numbers $\{\theta_i\}$. 
\end{theorem}
\begin{proof}

For $\gamma>(2-p)/(2p)$, we show that for $\varepsilon\in(0,1/2)$, the minimax code length $L(\varepsilon,\mathcal F)\notin \mathcal O(\varepsilon^{-1/\gamma})$ as $\varepsilon\to 0$. This can be done by contradiction. The main idea is to show that if we assume $L(\varepsilon,\mathcal F)\in \mathcal O(\varepsilon^{-1/\gamma})$, then the encoding length (as defined in Definition \ref{optimalrate}) will be too small to preserve enough information about functions in $\mathcal F$. The encoding-decoding process is thus unable to achieve error bound $\sup_{f\in\mathcal F}\|f-D(E(f))\|_{L^2}\leq \varepsilon$ \textit{\textbf{no matter how good}} the encoding-decoding process is. This can be treated as an analogue of a result from rate-distortion theory. 

Under the assumption, $\mathcal F$ has an embedded hypercube $\mathcal H$ of side $\delta$ (this can be chosen arbitrarily small) and dimension $m:=m(\delta)\geq C \delta^{-p}$ for some fixed $C>0$. Let $\xi=(\xi_1, \xi_2,\dots,\xi_m)\subset \{0,1\}^m$, then each vertex of the hypercube $H(m; f_0, (\psi_{i,m}))$ can be identified by 
$$h=f_0+\sum_{i=1}^m \xi_i \psi_{i,m}.$$

Suppose now we have a method of representing functions $f\in\mathcal F$ approximately by $R$ bits, that is, for all $f\in \mathcal F$ define $e:=\Enc(f)\in \{0,1\}^R$, and define a suitable decoding process maps $e$ to $\widetilde f$. 

Now we use the above process to encode the vertices $h$, then we get another point $\widetilde h$ which is not necessarily a vertex of hypercube. Let $\widehat h\in\mathcal H$ be the closest vertex to $\widetilde h$ in $L^2$ norm. Then we can assign a bitstring $\widehat\xi$ to $\widehat h$. Now the process starting from $\xi$ to $\widehat \xi$, can be described as $\widehat \xi = \Dec(\Enc(\xi))$. 

By orthogonality of hypercube, we have
\begin{align*}
    \|h-\widehat h\|_2^2 &= \delta^2 \sum_{i=1}^m (\xi_i-\widehat\xi_i)^2\\
    &=\delta^2 \text{Dist}(\xi, \widehat\xi).\\
    \therefore \max_{h\in\mathcal H} \{\|h-\widehat h\|_2^2 \} &\geq\text{Average}_{h\in\mathcal H} \{\|h-\widehat h\|_2^2\}\\
    &=\delta^2 \text{Average}_{h\in\mathcal H}(\text{Dist}(\xi, \widehat \xi))\\
    &\geq \delta^2 D_m(R)
\end{align*}
where the last result is from rate-distortion theory. 

For the sake of contradiction, we now fix $\gamma>(2-p)/(2p)$, then prove that for every pair of encoder-decoder $(E,D)\in\mathfrak E^\ell\times \mathfrak D^\ell$ with $\ell \in \mathcal O(\varepsilon^{-1/\gamma})$, it must result in the uniform error $\sup_{f\in\mathcal F} \|D(E(f))-f\|_{L^2}$ \textbf{not} bounded below by $\varepsilon$. Note that we should consider all kind of encoding process such that it chooses $n$ terms among the first $\pi(n)$ terms for $\pi$ a fixed polynomial. The fruitful result is that the error is not $\mathcal O(\varepsilon)$ for all kind of decoding process. 

Note that all $f\in \mathcal F$ has a corresponding basis representation 
$$f=\sum_{i=1}^\infty \theta_i\varphi_i.$$
Using polynomial depth-search encoding process, we choose $I_n\subset \{1,2,\dots,\pi(n)\}$ containing $n$ integers, then we choose the candidates $\{\theta_i\}_{i\in I_n}$, namely the sum $\sum_{i\in I_n} \theta_i\varphi_i$ to approximate $f$. However, this is not exactly how we encode $f$, as the coefficients $\{\theta_i\}_{i\in I_n}$ is not discrete. We see that $|\theta_i|\leq \|f\|_2\leq A$, hence we choose a number $\widetilde \theta_i$ from $[-A,A]\cap (\eta)\mathbb Z$ with $\eta=n^{-2/p}$ that is closest to $\theta_i$. Then the encoding-decoding process is symbolically defined by $D(E(f))=\widetilde f=\sum_{i\in I_n} \widetilde \theta_i \varphi_i$ as the coefficients $\widetilde\theta_i$ can be naturally decoded using the same idea. To explicitly write $E(f)$ as a bitstring, we first encode bitstrings of indices. There are $n$ indices, and each indices can be encoded by at most $\log_2\pi(n)=\mathcal O(\log n)$ bits, hence we have encoded $I_n$ using $\mathcal O(n\log n)$ bits. Then, note that $[-A,A]\cap (\eta)\mathbb Z$ contains at most $\lceil2An^{2/p}\rceil$ elements, we encode each coefficients using $\log_2\lceil2An^{2/p}\rceil=\mathcal O(\log n)$ bits, hence we can use $\mathcal O(n\log n)$ to encode the $n$ coefficients. Finally, we explicitly define $E(f)$ to be this bitstring of length $\ell$ with 
$$\ell\leq R(n):=\log_2\pi(n)+\log_2\lceil2An^{2/p}\rceil\leq C_1 n\log n.$$
($R(n)$ and $n\log n$ are actually equivalent)

Let $h\in \mathcal H$ be a vertex, using the encoding process above we have $\widetilde h$. Let $\widehat h\in \mathcal H$ be the closest vertex to $\widetilde h$. By rate-distortion theory, for $\rho<1/2$ if $n$ obeys $R(n)\leq \rho m$, then 
$$\max_{h\in\mathcal H} \{\|h-\widehat h\|_2^2\}\geq \delta^2 D_m(R(n))\geq \delta^2 D_1(\rho)m.$$

By construction, we have $\|\widetilde h - h\|_2\geq \|\widetilde h-\widehat h\|_2$, thus by triangle inequality we have
$$\|\widehat h-h\|_2\leq \|\widehat h-\widetilde h\|_2+\|\widetilde h-h\|_2\leq 2\|\widetilde h -h\|_2.$$

Now we refer to Definition \ref{optimalrate}. Estimation gives
\begin{align*}
    \sup_{f\in\mathcal F} \{\|f-D(E(f))\|_2^2\}&\geq \max_{h\in\mathcal H}\{\|h-\widetilde h\|_2^2\}\\
    &\geq \dfrac12 \max_{h\in\mathcal H}\{\|h-\widehat h\|_2^2\}\\
    &\geq \dfrac12\delta^2 D_1(\rho)m.
\end{align*}

Now assume $L(\varepsilon,\mathcal F)\in \mathcal O(\varepsilon^{-1/\gamma})$. Fix $\rho<1/2$, we choose $m:=m_k$ to be the smallest integer (see Definition \ref{hypercube}) and $\delta:=\delta_k$ so that $R(n)\leq C_1n\log n\leq \rho m$ and so $m\geq C\delta^{-p}$. From denseness of dimensions as we required in Definition \ref{hypercube}, we also have $\rho(m-1)<C_1n\log n\leq \rho m$. We then have $\delta^2m \geq m\cdot C^{2/p} m^{-2/p}=C^{2/p} m^{-(2-p)/p}\geq C'\left(\dfrac{n\log n+\rho}\rho\right)^{-\frac{2-p}p}$. Notice that if we only know $C_1 n\log n\leq \rho m$ then the last inequality will not be possible, hence it is important that $m$ can be any integer larger than a fixed large integer, not just being sufficiently large, which the latter could lead to $m$ being separated too much (compare Definition \ref{hypercube} and the original definition of hypercube in \cite{donoho2}).

This shows 
$$\sup_{f\in\mathcal F} \{\|f-D(E(f))\|_2\}\geq \sqrt{C'} (n\log n+\rho)^{-\frac{2-p}{2p}}\geq C'' \varepsilon^{\frac{2-p}{2p\gamma}}\quad\text{ for small }\varepsilon.$$
Now $\dfrac{2-p}{2p\gamma}<1$, hence the term $\varepsilon^{\frac{2-p}{2p\gamma}}$ is strictly larger than mere $\varepsilon$ when $\varepsilon\to 0$, contradicts to $\sup_{f\in\mathcal F}\{\|f-D(E(f))\|_2\}\leq \varepsilon$ (recall definition of $L(\varepsilon,\Star)$). We deduce that $L(\varepsilon,\mathcal F)\notin \mathcal O(\varepsilon^{-1/\gamma})$ for every $\gamma>\dfrac{2-p}{2p}$, hence $\gamma^*(\mathcal F)\leq \dfrac{2-p}{2p}$. 
\end{proof}

In the original proof, the author did not use the notion of optimal exponent, instead he used another quantity called the \quotes{optimal degree of sparsity}. It concerns the representation $f=\sum_{i=1}^\infty \theta_i \varphi_i$, whether one can tolerate only a few of $\theta_i$ being significantly large, while all the others being negligible, hence the word \enquote{sparsity}. Our proof above transfers result of optimal degree of sparsity to the optimal exponent.

Although not mentioned in \cite{donoho2}, it is important to emphasize that $\text{STAR}^\beta(C)$ does have an orthonormal basis $\{\varphi_i\}$, but they are not elements of the function class itself. 
Evidently, each element having $\|\cdot \|_{L^2}=1$ cannot be an element of $\text{STAR}^\beta(C)$. 
This is not a problem, as we only extract information about $\theta_i$s in the decomposition $f=\sum \theta_i \varphi_i$. We can also easily find an orthonormal basis of $L^2([0,1]^2)$. 
First note that $L^2([0,1]^2)$ is a separable linear space (\cite{brezis}, Theorem 4.13), hence having a countable dense subset $\{\alpha_i\}_{i\in\mathbb N}$.
Then, by using Gram-Schimdt process we can obtain an orthonormal basis $\{\varphi_i\}_{i\in\mathbb N}$. 
Any subspace of $L^2([0,1]^2)$ can then be represented as a countable linear combination of elements from this orthonormal basis. 

The last step is to show $\text{STAR}^\beta(C)$'s optimal exponent is bounded below by $\frac{2-p}{2p}$. Since this is a constructive proof, it is very detailed. We first list the result below.

\begin{theorem}\label{fourthstep}
The optimal exponent of $\Star$ obeys the lower bound $\gamma^*(\text{STAR}^\beta(C))\geq \dfrac\beta2$.
\end{theorem}

We introduce some notations, then we will directly apply Theorem 8.4 from \cite{donoho3}, modify them, in order to prove Theorem \ref{fourthstep} in this paper.

\begin{enumerate}[1.]
    \item \textbf{\textit{Edgels and Edgelets}}
    
    Let $m=2^k$ be some dyadic integer, then we define the \textbf{latticework} $\mathcal L(m)$ to be the union of all dyadic squares of edgelength $1/m$ in $[0,1]^2$ where their vertices are integer multiples of $1/m$.

Given two points $v_1,v_2\in[0,1]^2$, an \textbf{edgel} is the line segment $e=\overline{v_1v_2}$.

Given $n=2^J$, and for $m=2^j, 0\leq j\leq J$, we want to define a countable collection of edges connecting from the boundary of squares, because freely joining lines will resulted in an uncountable set of edges. Consider the collection of all dyadic vertices in $\mathcal L(n)$, which we have $(n+1)^2$ of them. Construct an edgel from each pair of vertices gives a collection of $\binom{(n+1)^2}2\in\mathcal O(n^4)$ elements, which is too big. We usually want a smaller collection. 

Let $\delta=2^{-J-K}$ be an even finer scale. At boundary of each dyadic square $S$ of length $1/m$, starting from upper-left vertex we pin down vertices $v_{i,S}$ in a clockwise fashion, with consecutive points distance as $\delta$. There are $M_j=4\times 2^{J+K-j}$ vertices on $S$. Let
$$E_\delta(S) = \{e=\overline{v_{i,S}v_{j,S}}, 0\leq i,j< M_j\},$$
then $E_\delta(S)$ contains $\binom{M_j}2$ edgels. We define the set of \textbf{edgelets} $\mathcal E(n,\delta)$ containing all possible edgels in some $E_\delta(S)$ for some $0\leq j\leq J$. The cardinality of $\mathcal E(n,\delta)$ is given as 
$$\#\mathcal E(n,\delta)=\sum_{j=0}^J 4^j\binom{M_j}2\leq \sum_{j=0}^J 4^j\dfrac{M_j^2}2=\sum_{j=0}^J 4^j\cdot 8\cdot 2^{2J+2K-2j}=8(J+1)\delta^{-2}.$$
In the case where $\delta=2^{-J}=1/n$, then $\#\mathcal E(n,\delta)\leq 8(\log_2n+1)n^2\in \mathcal O(n^2\log n)$. Hence we can use roughly $\mathcal O(n^2)$ elements to serve as a basis for edgels approximation. Although each edgelet in $\mathcal E(n,\delta)$ only contained in a small dyadic square, but their collection is manageable, they express many rotation, location and anisotropic features. On the other hand, we can chain multiple edgelets together to compensate their ability to approximate longer edgels. 

\item \textbf{\textit{Recursive Dyadic Partition (RDP)}}

This is a discussion about an adaptive approach used to approximate functions in $\Star$. Generally, a simple closed curve in $[0,1]^2$ might have some very smooth part, and some ill-looking part. Therefore, we hope to using the least resource to approximate them: Using large dyadic squares for the smooth parts, while using finer scale dyadic squares to handle the part with more perturbation. 

We define \textbf{Recursive Dyadic Partition (RDP)}. The trivial RDP is $\{[0,1]^2\}$. Now if $\mathcal P$ is an existing RDP, then we choose a square $S\in \mathcal P$, perform a quad-split so it is partition into four smaller dyadic squares (called children). A finer RDP can be construct by simply adding these four squares into $\mathcal P$ and remove the original parent. We can also construct a coarser RDP from an existing one: by merge four adjacent dyadic squares into one larger square (called ancestor), add the larger square and remove four of them from the RDP. We also call four adjacent dyadic squares being siblings.   

We can think of RDP as a quadtree $Q$, the bottom-most squares are those elements in the RDP, and in between the tree are ancestors of the elements in the RDP.

Now we involve the concept of edgelets into RDP. For $n=2^J, \delta=2^{J+K}$, we define \textbf{Edgelet-decorated RDP (ED-RDP)} to be an RDP $\mathcal P$ where each $S\in\mathcal P$ can be splitted by an edgelet in $\mathcal E(n,\delta)$. A typical element of an ED-RDP is either a dyadic square (unsplit) or a wedgelet resulted from split square. We also assume that if a square is split into two wedgelets, then the wedgelets cannot be split again. Since each square in an ED-RDP can only be split into at most two wedgelets, if $w$ is a wedgelet already in $\overline{\mathcal P}$, then no other wedgelet can be contained in the same square where $w$ live in. It is because if two wedgelets in $\overline{\mathcal P}$ form a square, then there is no point in splitting the square beforehand. We denote the collection of all ED-RDPs as ED-RDP$(n,\delta)$. 

\item \textbf{Averages as $n\times n$ array}

In an ED-RDP$(n,\delta)$, $\overline{\mathcal P}$, we choose a square/wedgelet $P\in \overline{\mathcal P}$ and let $1_P$ be its characteristic function on $[0,1]^2$. Now we define $\widetilde{1_P}$ be an $n\times n$ array such that its $(i,j)$ entry is equal to the average of $1_P$ on the square $S_{i,j}=[(j-1)/n,j/n]\times [(i-1)/n,i/n]$. Formally, 
$$(\widetilde{1_P})_{i,j}=\dfrac1{1/n^2}\int_{S_{i,j}} 1_P,$$
and we also define $L(\overline{\mathcal P})$ to be the linear space of $n\times n$ arrays spanned by the collection $\{1_P\}_{P\in\overline{\mathcal P}}$. 
We can also regard $\widetilde 1_P$ as a function on $[0,1]^2$, which takes $n^2$ (possibly) different values on each dyadic square, then $L(\overline{\mathcal P})$ can be regarded as a subspace of functions in $L^2([0,1]^2)$. 
By construction of $\overline{\mathcal P}$, we know that two different member $P_1,P_2$ have disjoint interior, and if $S$ is a dyadic square of side-length $1/n$ and intersect $P_1$ in interior, then $S$ must be disjoint from $P_2$. 
This shows two different $n\times n$ arrays $\widetilde{1_{P_1}}, \widetilde{1_{P_2}}$ are orthogonal, with inner product defined by $(f,g)=\int_{[0,1]^2}fg$. 

For $f\in\Star$ defined on $[0,1]^2$ and $P\in\overline{\mathcal P}$, we define the projection coefficient by the integral $$f_P:=\int_{[0,1]^2}\dfrac{f\widetilde{1_P}}{(\widetilde{1_P},\widetilde{1_P})}$$
Then we define 
$$\ave(f| \overline{\mathcal P})=\sum_{P\in\overline{\mathcal P}} f_P \widetilde{1_P}$$
which is the least-square projection of $f$ to the linear space $L(\overline{\mathcal P})$. Again, we may regard $\ave(f|\overline{\mathcal P})$ as an array or as a function on $[0,1]^2$. 
\end{enumerate}

Donoho proved the following theorem, noted as Lemma 8.4 in his paper \cite{donoho3}.

\begin{theorem}\label{donohomain}
Let $1<\beta\leq 2, 0<C<\infty$, $n=2^J, \delta=2^{J+K}$. For each $f\in \Star$ there exists a corresponding RDP $\mathcal P_f$ with fewer than $n'=K_1\cdot n+K_2$ elements, and a corresponding ED-RDP$(n,\delta)$, denoted as $\overline{\mathcal P}_f$ such that 
$$\|f-\ave(f|\overline{\mathcal P}_f)\|_{L^2}^2\leq K_\beta C_\beta n^{-\beta}+\delta,$$
where $K_1,K_2$ are constants and $K_\beta$ depends on $\beta$ only. 
\end{theorem}

We now apply this result to prove Theorem \ref{fourthstep}, with some modification.

\begin{proof}[Proof for Theorem \ref{fourthstep}]
The optimal exponent is based on an encoding-decoding process which encode the function into a bitstring. Therefore, we want to discover a countable dictionary based on $\{1_P\}_{P\in \overline{\mathcal P}}$, take their finite linear combination based on polynomial search-depth, then discretize their coefficients to form an approximation. 

The approximant $\ave(f|\overline{\mathcal P_f})$ is a linear combination of $\{\widetilde 1_P\}_{P\in\overline{\mathcal P}_f}$. This is an orthogonal system as noted before, we can change them into orthonormal system $\{\phi_P\}_{P\in\overline{\mathcal P}_f}$ by define
$$\phi_P = \dfrac{\widetilde 1_P}{\sqrt{(\widetilde 1_P, \widetilde 1_P)}}=\dfrac{\widetilde 1_P}{\sqrt{\int_{[0,1]^2}(\widetilde 1_P)^2}},$$
and it is convenient to use the notation $\|\widetilde 1_P\|_{L^2} = \sqrt{\int_{[0,1]^2} (\widetilde 1_P)^2}$. Then we have the below orthonormal presentation
$$\ave(f|\overline{\mathcal P}_f)=\sum_{P\in \overline{\mathcal P}_f}f_P\|\widetilde 1_P\|_{L^2} \phi_P.$$
By the classical Cauchy-Schwarz inequality, we have
$$\left|f_P\|\widetilde 1_P\|_{L^2}\right|=\dfrac1{\|\widetilde 1_P\|_{L^2}} \left|\int_{[0,1]^2} f\widetilde 1_P\right|\leq \|f\|_{L^2}\leq 1.$$

Now we choose $\eta=n^{-2}$, then we define $\theta_P$ to be an integer multiple of $\eta$ which is closest to $f_P\|\widetilde 1_P\|_{L^2}$, if there is a tie, we define $\theta_P$ to be closer to zero. This discretization makes $\theta_P\in[-1-\eta,1+\eta]\cap \eta\mathbb Z$, which the latter set has at most $3\eta^{-1}$ elements when $\eta$ is small enough. We define the discrete, slightly distorted approximant of $f$ as 
$$\widetilde f = \sum_{P\in \overline{\mathcal P}_f}\theta_P \phi_P.$$

Then we find that when $J=K$, that is $\delta=1/n^2$, we have
\begin{align*}
    \|f-\widetilde f\|_{L^2}&\leq \|f-\ave(f|\overline{\mathcal P}_f)\|_{L^2}+\|\ave(f|\overline{\mathcal P}_f)-\widetilde f\|_{L^2}\\
    &\leq \sqrt{K_\beta C_\beta n^{-\beta}+\delta}+\sum_{P\in \overline{\mathcal P}_f} |\theta_P-f_P\|\widetilde 1_P\|_{L^2}|\\
    &\leq K_\beta^{1/2}C_\beta^{1/2}n^{-\beta/2} + \delta^{1/2} +\frac12 (\# \overline{\mathcal P}_f)\eta\\
    &\leq K_\beta^{1/2}C_\beta^{1/2}n^{-\beta/2} + n^{-1} + K_3\cdot n^{-1}
\end{align*}
where $K_3$ is some constant. Notice that the term with $n^{-\beta/2}$ dominate the whole expression.

Now we fix $\gamma<\beta/2$. Given $\varepsilon>0$, we exhibit a pair $(E_\ell,D_\ell)$ of encoding-decoding process while encoding length $\ell$ obeys the order $\mathcal O(\varepsilon^{-1/\gamma})$ and the error across all element of $f\in \Star$ satisfies $$\sup_{f\in\Star}\|f-D_\ell(E_\ell(f))\|_{L^2}\leq \varepsilon.$$
Then this will proves $L(\varepsilon,\Star)\leq \inf_{(E,D)}\ell\in \mathcal O(\varepsilon^{-1/\gamma})$. All we have now is $\varepsilon>0$ is fixed. We choose a smallest positive integer $J$ such that $n=2^J$ satisfies 
$$(K_\beta C_\beta)^{1/2} n^{-\beta/2}+(1+K_3)n^{-1}\leq \varepsilon.$$
This implies the above inequality fails when one substitutes $n$ by $n/2$. Thus
\begin{align*}
    (2^{\beta/2}(K_\beta C_\beta)^{1/2}+2(1+K_3))n^{-\beta/2}&\geq (K_\beta C_\beta)^{1/2} \left(\frac n2\right)^{-\beta/2}+(1+K_3)\left(\dfrac n2\right)^{-1}\\
    &>\varepsilon,
\end{align*}
which means there is a constant $Q(\beta,C)$ only depends on $\beta$ and $C$ such that 
$$n\leq Q(\beta,C)\varepsilon^{-2/\beta}.$$

Next, we describe an encoding process for exactly record $\widetilde f$. Note that solely from $\varepsilon$ we know how to choose $J$. Thus we encode $J$ into bitstring, which uses $\log_2J=\log_2\log_2n$ length. 
To record which orthonormal elements $\phi_P$ are used from the $\overline{\mathcal P}_f\in \text{ED-RDP}(n,\delta)$, recall the collection of all edgelets of dyadic squares with side-length $\geq 1/n$, drawn at $\delta=1/n^2$ equally spaced level has cardinality
$$\#\mathcal E(n,\delta)\leq 8(J+1)\delta^{-2}\leq 8(\log_2n+1)n^4\leq 16n^5.$$
This is also our required polynomial search depth, we may index each element of $\mathcal E(n,\delta)$ up to $16n^5$. Moreover, we only choose at most $\# \overline{\mathcal P}_f\leq  K_3\cdot n$ elements to approxmiate $f$, each index require $\log_2 16n^5$ bits to encode. Therefore, we record positions of elements by encode their index in order, using at most $K_3\cdot n\log_2(16n^5)$ bits. Then, we record the coefficient $\theta_P$ of each chosen orthonormal element $\phi_P$. Recall each coefficient can be encoded using $\log_2(3\eta^{-1})$ bits, hence we use at most $K_3\cdot n\log_2(3\eta^{-1})=K_3\cdot n\log_2(3n^2)$ bits to encode all coefficients. It is clear from now that we can decode this bitstring back to $\widetilde f$ in the natural way. We conclude that the above process describe an encoding-decoding pair $(E_\ell,D_\ell)$ such that $D_\ell(E_\ell(f))=\widetilde f$ for all $f\in\Star$ where the encoder length obeys
$$\ell \leq K_3\cdot n\log_2(16n^5)+K_3\cdot n\log_2(3n^2) + \log_2\log_2n\leq K_4 n\log n $$
for some constant $K_4$ independent from $\beta$ and $C$. Now we find that 
\begin{align*}L(\varepsilon,\Star)\leq K_4n\log n&\leq K_4 Q(\beta,C_\beta)\varepsilon^{-2/\beta} \log(Q(\beta,C_\beta)\varepsilon^{-2/\beta})\in \mathcal O(\varepsilon^{-1/\gamma}).\end{align*}
This completes the proof. 
\end{proof}

\section{Conclusion}
In this paper, we focus on the theoretical aspect of neural networks, which we seek the problem of neural network approximation by emphasizing their actual limitation, rather than using learning algorithms such as stochastic gradient descent and adaptive moment estimation. Furthermore, we found that meaningful discussion of approximation quality by neural networks can appear if we introduce some limitation to the settings of neural networks. 
This leads us to the effective best $M$-term/$M$-edge approximation rates. 
To link neural networks to what we are familiar with, such as the Fourier series, we define (effective) representability of representation systems by neural networks, then transfer existing knowledge about representation systems to neural networks. This transition makes neural networks become an extremely powerful tool in modern technology.

Finally, we introduce two applications by using the previous settings and theoretical results. One is approximate B-spline functions by neural networks, while the former is used to generate B-spline curves. We have shown that the class of B-spline functions is effectively representable by neural networks. This enable us to use neural networks to reconstruct B-spline curves with an acceptable distortion. The other example is to apply rate-distortion theory and discrete wedgelets construction to prove the class of $\beta$ cartoon-like functions has a finite optimal exponent. These two practical adaptations of theories showed that it is possible for us to combine the theoretical and practical aspects of neural networks in the future, implying examine the theoretical limitation of gradient descent, adaptive moment estimation and other learning algorithm might be possible after all. Therefore, the results stated in this paper can be used to pursue the extreme limitation of neural networks.

\bibliography{main}  
\bibliographystyle{apalike}

\end{document}